\pdfoutput=1

\documentclass[11pt]{article}

\usepackage[]{acl}

\usepackage{times}
\usepackage{latexsym}

\usepackage{graphicx}
\usepackage{multirow}
\usepackage{color,soul}

\usepackage[T1]{fontenc}

\usepackage[utf8]{inputenc}

\usepackage{microtype}

\title{Faithful Low-Resource Data-to-Text Generation through Cycle Training}

\author{Zhuoer Wang$^{\dagger1}$\quad Marcus Collins$^{\star2}$\quad Nikhita Vedula$^{\star2}$\\
\textbf{Simone Filice$^2$\quad Shervin Malmasi$^2$\quad Oleg Rokhlenko$^2$} \\
\\
$^1$Texas A\&M University \quad $^2$Amazon\\
\texttt{wang@tamu.edu} \\
\texttt{\{collmr,veduln,filicesf,malmasi,olegro\}@amazon.com}
}

\begin{document}
\maketitle

\renewcommand{\arraystretch}{1.2}

\renewcommand{\thefootnote}{\fnsymbol{footnote}}
\footnotetext[2]{The research was done during an internship at Amazon.}
\footnotetext[1]{These two authors contributed equally to this work.}
\renewcommand{\thefootnote}{\arabic{footnote}}

\begin{abstract}
Methods to generate text from structured data have advanced significantly in recent years, primarily due to fine-tuning of pre-trained language models on large datasets. However, such models can fail to produce output faithful to the input data, particularly on out-of-domain data. Sufficient annotated data is often not available for specific domains, leading us to seek an unsupervised approach to improve the faithfulness of output text. Since the problem is fundamentally one of consistency between the representations of the structured data and text, we evaluate the effectiveness of \textit{cycle training} in this work. Cycle training uses two models which are inverses of each other: one that generates text from structured data, and one which generates the structured data from natural language text. We show that cycle training, when initialized with a small amount of supervised data (100 samples in our case), achieves nearly the same performance as fully supervised approaches for the data-to-text generation task on the WebNLG, E2E, WTQ, and WSQL datasets. We perform extensive empirical analysis with automated evaluation metrics and a newly designed human evaluation schema to reveal different cycle training strategies' effectiveness of reducing various types of generation errors.
Our code is publicly available at \url{https://github.com/Edillower/CycleNLG}.

\end{abstract}

\section{Introduction}
\label{sec:intro}
A wealth of information exists in the form of structured knowledge, such as movie information databases or product catalogs, which we may want to verbalize for a variety of purposes, such as comparing two items, or presenting detailed descriptions in a natural language form suitable for conversational assistants.
Recent work has tackled this data-to-text generation task using freely available public datasets, most notably WebNLG \cite{WebNLG2020report} and ToTTo \cite{ToTTo}. However, there remain two major challenges. 
First, the volume of training data required for good performance, especially if it is not in a domain represented by one of the existing corpora, is very large.
Second, multiple recent papers \cite{TableFormer, ToTTo}, \textit{inter alia}, point out that neural natural language generation (NLG) from structured data tends to produce multiple kinds of errors which limit the utility of these models in customer-facing applications. Hallucinations occur when NLG models inject nonsensical words or information not related to the input structured data, into the generated output text. For instance, an NLG model may claim a shirt's color is ``three''. Simple factual errors occur when an NLG model produces coherent but factually wrong output.

There are two threads of research to consider as we attempt to tackle these problems in the data-to-text setting. The first is designing models that directly produce output more faithful to the input data. The second is designing models to detect and correct factual errors or hallucinations after the output text is generated.
In both cases, prior research has generally assumed sufficient pairs of structured data and text as training data to achieve human-level performance on the task. While fact verification models can achieve very high performance, they generally do so when trained on large corpora of 100,000 examples or more. Since performance appears to degrade when evaluated on out-of-domain data~\cite{vedula2022emnlp}, this presents a significant limitation of fact-verification models. Similarly, corpora like WebNLG contain about 20,000 examples; this is probably too small to achieve human performance even under full supervision \cite{CycleGT} but is large enough to make it prohibitive to generate domain-specific corpora of the size of WebNLG.

In spite of the above mentioned limitations, very few of the models developed for data-to-text and table-to-text tasks take advantage of the fact that the task of faithful text generation is fundamentally one of \textit{consistency} between the data and the corresponding text. In fact, despite the WebNLG 2020 challenge being explicitly bi-directional, only three models competing in the challenge leveraged this idea of consistency. 

To overcome the aforementioned limitations related to the lack of training data (especially out-of-domain data) and the consistency between structured data and text, we adopt a Cycle Training~\cite{CycleNER} approach. We assume \textit{unpaired} data $\mathcal{D}$, in the form of subject-predicate-object triples, and text $\mathcal{T}$, which may or may not be from the same domain. We also make use of a small (100 samples) set of paired data and text, $\mathcal{D}_{pr}, \mathcal{T}_{pr}$. Cycle training makes use of two iteratively trained models, a forward model $\mathcal{F}:\mathcal{D}\rightarrow\mathcal{T}$ and a reverse model $\mathcal{R}:\mathcal{T}\rightarrow\mathcal{D}$. Training is unsupervised, namely, we freeze one model and use it to transform one set of inputs, and train the other by using it to predict the original input from the output of the first model. Concretely, in one cycle, we freeze $\mathcal{F}$, and train $\mathcal{R}$ by reconstructing the input $\mathcal{D}$ as $\mathcal{R}(\mathcal{F}(\mathcal{D}))$. After one training epoch, we reverse the roles of the two models. Remarkably, even though the models are initially quite poor, this can converge to models with near-supervised performance, as we will show. Moreover, we show that this process ensures the \textit{faithfulness} of the output text with respect to the input data, and vice versa, even with very little or no paired data.
 
We note that a previous data-to-text system, CycleGT, has used cycle training \cite{CycleGT}. We will discuss in detail the differences between CycleGT and our proposed approach in Section \ref{related}. Moreover, we examine in detail the conditions under which cycle training works well, with an emphasis on domains and the nature of the training text and structured data. We find that unsupervised cycle training outperforms low-resource fine-tuned models and can achieve near fully-supervised performance when initialized and post-tuned with a small amount of annotated data. We detail the results and findings in Section \ref{sec:results-n-discussion}.
Thus, to build on past research in self-consistent data-to-text generation, we make these novel contributions:

(i) We successfully apply cycle training to both the data-to-text and text-to-data models using only a pre-trained language model, T5, without recourse to graph methods or other auxiliary models.

(ii) We show that cycle training achieves nearly the same performance as supervised models for some domains. 

(iii) We present an extensive empirical analysis on the conditions under which cycle training works well, and on the data-to-text faithfulness with respect to different types of generation errors.

(iv) We design a novel counting and ranking based annotation schema to more comprehensively evaluate the faithfulness of the generated text from the standpoints of correctness, faithfulness, data coverage, and fluency. Our schema improves upon the rating-based schema used for the WebNLG 2020 Challenge, in terms of objectiveness, consistency, precision and ease of evaluation.

\section{Related Work}
\label{related}
Multiple data-to-text and table-to-text tasks have been presented in the literature, such as WebNLG \cite{WebNLG-creation, WebNLG-DBPedia, WebNLG-RDF}, DART \cite{DART}, ToTTo \cite{ToTTo}, and WikiTableT \cite{WikiTableT}, which primarily consist of data from general-purpose sources like Wikipedia. Several large language models~\cite{TAPAS,TAPEX,TableFormer} have been trained on large scale table-to-text corpora~\cite{2019TabFactA} to perform fact verification. 
However, these models may not perform well on specific domains they have not been trained on, such as e-commerce~\cite{vedula2022emnlp,vedula2022matters}. 
Therefore, we must either find a way to easily generate new data to train large data-to-text models, or use unsupervised methods. 
Recently, \citet{xiang-etal-2022-asdot} attempted to augment training data using GPT-3 \cite{NEURIPS2020_1457c0d6},  and \citet{su-etal-2021-shot-table} employed an information retrieval system to build prototypes for the generation. Our work makes orthogonal contributions to these studies, as we directly utilize the underlying unpaired data and text of a target corpus without recourse to any additional information retrieval or generation systems.
Further, the above-mentioned data-to-text tasks have been evaluated primarily on automatic word- or n-gram-level metrics such as BLEU \cite{papineni-etal-2002-bleu} or METEOR \cite{banerjee-lavie-2005-meteor}, with minimal (and mostly subjective) evaluation of faithfulness. 
In this work, we design a novel annotation schema to perform a more comprehensive evaluation of the faithfulness of the generated text to the input data. 

Cycle training~\cite{UnpairedImageImage, 10.1109/cvpr.2016.20} relies on two models which are essentially inverse transforms of each other that are used to create ``cycles'', which should return identical output to the input given. 
There are two distinct forms of cycle training. The first form \cite{10.1109/cvpr.2016.20} aims to learn to transform from one input form to another, e.g., to learn rotations of a car in one image to another. 
The second is the use of a ``cycle consistency loss'' as an auxiliary loss to some other task, e.g., in generative adversarial networks performing style transfer on images~\cite{UnpairedImageImage}. 
NLG typically relies on models which are auto-regressive and non-differentiable. This precludes the direct use of cycle consistency losses \cite{CycleGT,NonParaTextTransfer,CycleNER}. Nonetheless, we can still use cycle training via an alternating training strategy where we freeze one model and train the other, and vice versa~\cite{UnsuperNMT,NonParaTextTransfer}. In this work, we train solely using cycle consistency. 
Cycle training has been recently applied to language processing tasks. In one text-to-text application, \citet{iovine-etal-2022-cyclekqr} use a similar unsupervised methodology to perform bidirectional text transformations for converting keyword search queries to natural language questions, and \textit{vice versa}.
It has also been used for Named Entity Recognition in the absence of large annotated text \cite{CycleNER}. In this case, one model extracts entities, and the inverse model creates text from those entities. The approach is limited by the fact that there are many ways to realize sentences with the same entities. Put differently, there is no strong requirement of cycle consistency, and this will become even more apparent as we analyze the conditions under which cycle training works well in data-to-text tasks.  

To the best of our knowledge, the only work to explicitly call out the self-consistency requirement of data-to-text generation tasks is the CycleGT model~\cite{CycleGT} developed for data-to-text generation on the WebNLG dataset. 
One key advantage of cycle training is that it need not rely on any supervision, and instead relies primarily or solely on the self-consistency of inputs and outputs. However, CycleGT relies on a pre-existing NER model to extract entities from the output text. The authors then train an inverse model to predict the links between entities and predicates. Should the entities not be recognized by their NER system, the model will fail overall; this is not an uncommon situation in applications such as online shopping \cite{vedula2022emnlp,vedula2023generating}, where entities are complex or change frequently \cite{malmasi-etal-2022-semeval}. In principle, a separate NER model could be built using cycle training, as in CycleNER \cite{CycleNER}, but the CycleGT authors did not do so. In this work, we design a simple approach using pre-trained language generation models, fine-tuned for both data-to-text and text-to-data generation cycles. 

\section{Methodology}
\label{sec:methods}

\begin{figure*}[t]
\centering
\includegraphics[width=0.8\textwidth]{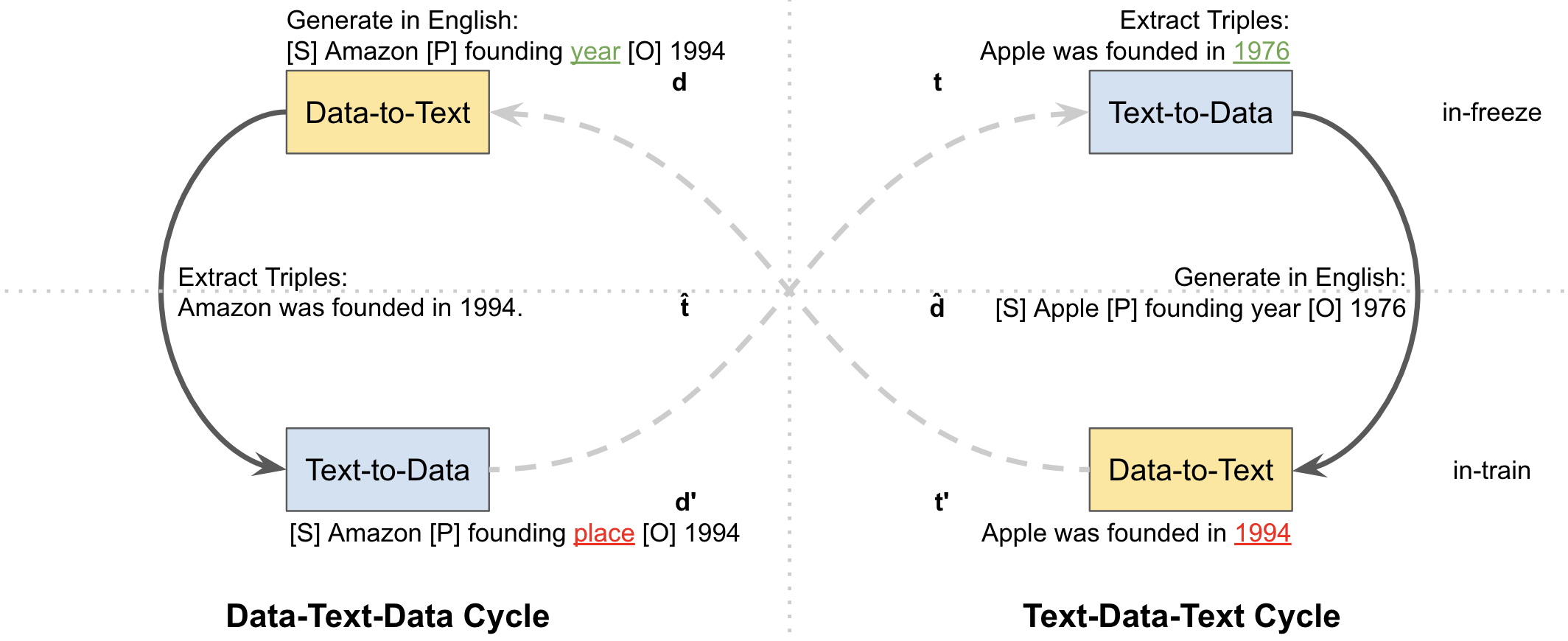}
\vspace{-5pt}
\caption{Cycle Training of the Data-to-Text model and Text-to-Data model. For each cycle, the upper-level models are frozen to generate the intermediate text for the training of the lower-level models, that attempt to reconstruct the initial inputs (\textit{d}, \textit{t} denote initial inputs of the upper-level models; \textbf{$\hat{t}$}, \textbf{$\hat{d}$} denote the upper-level models' generations that serve as inputs to the lower-level models; \textit{d'}, \textit{t'} denote the generations of the lower-level models).}
\label{fig1}
\end{figure*}

\subsection{Backbone Models}
\label{subsec:backbone-models}
The pre-requisite of cycle training is having two mutually inverse models. We adopt T5, an evidently strong-performing model according to the WebNLG 2020 challenge \cite{WebNLG2020report,agarwal-etal-2020-machine,CycleGT}, as our backbone model for both forward generation, (\begin{math}\mathcal{F}: \mathcal{D} \rightarrow \mathcal{T}\end{math} that performs RDF-to-text generation) and reverse generation, (\begin{math} \mathcal{R}: \mathcal{T} \rightarrow \mathcal{D}\end{math} that performs text-to-RDF generation). T5 is a large sequence-to-sequence model pre-trained with the unsupervised span-mask denoising objective and several supervised text generation tasks like summarization and translation \cite{t5}. We linearize the RDF triples of each sample into a sequence \begin{math}d\end{math} that denotes the subject, predicate, and object of each triple by the \texttt{[S]}, \texttt{[P]}, and \texttt{[O]} tags respectively. Therefore, both RDF-to-text and text-to-RDF can be treated and trained as sequence-to-sequence generation tasks. 
We further train or optionally fine-tune the T5 backbone models, as detailed in Section \ref{sec:experiments}, with the teacher forcing \cite{williams1989learning,lamb2016professor} learning objective for task-specific generation. This means that for the training of the auto-regressive decoder, we do not propagate the model decoded next token but force each input to be the correct gold token for training.  

\subsection{Cycle Training of the Backbone Models}
\label{subsec:ct}

Iterative Back-Translation (IBT) \cite{hoang2018iterative} has been reported as an effective training schema that enforces cycle consistency for various NLP tasks \cite{CycleGT,CycleNER}. We apply this idea to iteratively cycle train our models. This consists of the Data-Text-Data (\textbf{DTD}) cycle that enforces the self-consistency of data, and the Text-Data-Text (\textbf{TDT}) cycle that similarly enforces the self-consistency of text. 
As shown in Figure \ref{fig1}, for the DTD cycle, the Data-to-Text model takes the linearized triples \begin{math}d\end{math} as input and generates the associated intermediate text \begin{math}\hat{t}\end{math}. Sequentially, the Text-to-Data model is trained with the objective of reconstructing \begin{math}d\end{math} with the supplied \begin{math}\hat{t}\end{math}. The reconstruction loss \begin{math}\mathcal{L}_{d'}\end{math} is the averaged negative log likelihood shown below where \begin{math}d_{i}\end{math} denotes the \begin{math}i\end{math}-th token of sequence \begin{math}t\end{math} and \begin{math}|d|\end{math} is the sequence length:

\begin{center}
\begin{math}\mathcal{L}_{d'} = - \frac{1}{|d|}\sum_{i=0}^{|d|}\log p(d_{i}|d_{0},...,d_{i-1},\hat{t}) \end{math}
\end{center}

\noindent In a reverse manner, for the TDT cycle, the Text-to-Data model first takes text \begin{math}t\end{math} as input and generates the associated linearized triples \begin{math}\hat{d}\end{math}. Sequentially, the Text-to-Data model is trained with the objective of reconstructing \begin{math}t\end{math} with the supplied \begin{math}\hat{d}\end{math}. The reconstruction loss \begin{math}\mathcal{L}_{t'}\end{math} is the averaged negative log likelihood shown below where \begin{math}t_{i}\end{math} denotes the \begin{math}i\end{math}-th token of sequence \begin{math}t\end{math} and \begin{math}|t|\end{math} is the sequence length:

\begin{center}
\begin{math}\mathcal{L}_{t'} = - \frac{1}{|t|}\sum_{i=0}^{|t|}\log p(t_{i}|t_{0},...,t_{i-1},\hat{d})\end{math}
\end{center}

Due to the non-differentiable procedure of generating discrete intermediate outputs of tokens, the reconstruction loss can only propagate through the second model of each cycle, namely the Text-to-Data model of the DTD cycle and the Data-to-Text model of the TDT cycle. Therefore, the training of the two models can only proceed with the alternation of the TDT cycle and the DTD cycle so that both models' performance may gradually improve.

\section{Experimental Setup}
\label{sec:experiments}

\subsection{Data and Baselines}
\label{subsec:data}

\begin{table*}[t]
\footnotesize
\centering
\begin{tabular}{l|l|c|r|c|r|c}
\textbf{Dataset} &
  \textbf{Domain} &
  \textbf{\begin{tabular}[c]{@{}c@{}}Split Size\\ (Train/Dev/Test)\end{tabular}} &
  \textbf{\begin{tabular}[c]{@{}c@{}}Unique\\ Predicates\end{tabular}} &
  \textbf{\begin{tabular}[c]{@{}c@{}}Triples/Sample\\ (median/max)\end{tabular}} &
  \textbf{\begin{tabular}[c]{@{}c@{}}Vocab\\ Size\end{tabular}} &
  \textbf{\begin{tabular}[c]{@{}c@{}}Tokens/Sample\\ (median/max)\end{tabular}} \\ \hline
\textbf{WebNLG} & DBPedia (16 categories) & 35,426/4,464/7,305 & 1,236 & 3 / 7  & 20,126 & 21 / 80  \\ 
\textbf{E2E}    & Restaurants            & 33,482/1,475/1,475 & 41    & 4 / 7  & 6,158  & 22 / 73  \\ 
\textbf{WTQ}    & Wikipedia (open-domain) & 3,253/361/155      & 5,013 & 2 / 10 & 11,490 & 13 / 107 \\ 
\textbf{WSQL}   & Wikipedia (open-domain) & 526/59/38          & 946   & 2 / 6  & 2,353  & 12 / 34  \\
\end{tabular}
\caption{Datasets statistics and comparison.}
\label{table0}
\end{table*}

We experiment on existing data sources that have annotated pairs of data triples and reference texts.

\noindent\textbf{WebNLG} \cite{WebNLG-DBPedia, WebNLG-RDF, WebNLG2020report} is a well-established dataset that has supported multiple challenges on four tasks: RDF-to-English (Text), RDF-to-Russian (Text), English (Text)-to-RDF, and Russian (Text)-to-RDF. Each WebNLG sample consists of a set of subject-predicate-object triples and up to three associated human-written reference texts that faithfully express and verbalize the information contained in the triple set. We use the English data from the most recent 3.0 version of the WebNLG corpus, from the WebNLG+ 2020 challenge. 

\noindent\textbf{DART} \cite{DART} is a large-scale data-to-text dataset that unifies and builds upon multiple data resources including E2E \cite{novikova-etal-2017-e2e}, WikiSQL (WSQL) \cite{zhong2017seq2sql}, WikiTableQuestions (WTQ) \cite{pasupat-liang-2015-compositional}, and WebNLG \cite{WebNLG-creation}. To better facilitate our experiments and evaluations on different domains, we separately utilize the human-annotated portion of E2E, WTQ, and WSQL from DART. To align the data formats in accordance with WebNLG, we also drop some WSQL and WTQ samples that contain non-conventional structural tags. The DART dataset hereafter refers to the cleaned, WebNLG-excluded, and human-annotated portion of E2E, WTQ, and WSQL.

Table \ref{table0} shows detailed dataset statistics. When the data is used for cycle training, we follow previous work and split all the paired samples into one separate corpus of shuffled text, and another separate corpus of shuffled triple sets. For the linearized sequences, as shown in Figure \ref{fig1}, we: (1) prefix the string {\small``{\texttt{Generate in English:}}''} to the input sequence of the RDF-to-text model and prefix the string {\small``{\texttt{Extract Triples:}}''} to the input of the text-to-RDF model; (2) convert camel-cased or snake-cased subjects, predicates and objects to regular strings; and (3) normalize accented characters.

\label{subsec:baseline}

Fine-tuning large pre-trained language models, such as BERT \cite{devlin-etal-2019-bert}, BART \cite{lewis-etal-2020-bart}, and T5 \cite{t5} has been proven to be effective in achieving new state-of-the-art performance on numerous tasks. Fine-tuning refers to the supplemental training of a pre-trained model on a dataset of the target task and domain. We detail and perform the following three baseline fine-tuning strategies in this work:

\smallskip

\noindent\textbf{Fully supervised fine-tuning:} We fine-tune T5 with the entire in-domain (with respect to the test set) data as the supervised baseline. 

\noindent\textbf{Low-resource fine-tuning:} We fine-tune the T5-base model with 100 randomly selected sets of triples and their associated reference texts to formalize a low-resource supervised baseline. We deem 100 annotated samples to be a small enough amount, that is easily achievable with a relatively low human annotation effort.

\noindent\textbf{Low-resource fine-tuning with additional pre-training:} When using text from the target domain for cycle training, the teacher forcing algorithm naturally raises the probability of generating the target domain tokens, which may result in performance gains in token matching metrics (Section \ref{subsec:automatic-eval}). To study the influence of using in-domain text, we further pre-train the T5 model with in-domain text and an unsupervised span-mask denoising objective prior to the low-resource fine-tuning process. 
As our main objective is to probe a training strategy orthogonal to the model structure, we only include the above three baselines to control the model structure, data pre-requisites, and parameter sizes. 

\subsection{Comparing Cycle Training Strategies and Pre-requisites}
\label{subsec:ct-strategies}

We explore two different training strategies evaluating the effectiveness and generalizability of cycle training under different data constraints.

\noindent\textbf{Unsupervised cycle training:} As the most constrained low-resource scenario, in unsupervised cycle training we directly employ the IBT schema to cycle-train the forward model and reverse model with unpaired text and triple sets in turns. 

\noindent\textbf{Low-resource cycle training:} In this setting, a small amount of paired text and triple sets are accessible. For fair comparison and consistency, we utilize the same subset of data as the low-resource fine-tuning baseline described in Section \ref{subsec:data}. The low-resource paired data is leveraged through \emph{pre-cycle fine-tuning}, which first trains the forward and reverse model with the paired data before employing the IBT schema to cycle-train the two models. 

\citet{CycleGT} and \citet{CycleNER} vaguely state that the latent content or entity distribution of the text corpus and the data corpus must have some uncertain degree of overlap to make the cycle training approach work. To empirically assess this pre-requisite condition, we apply unsupervised cycle training with the same size of text and data corpus at different matching levels, as a rough approximation of overlap of the latent content or entity distribution. Specifically, we randomly select half of the WebNLG triplets as the data corpus. We purposefully select five equal-sized text corpora that contain 0\%, 25\%, 50\%, 75\%, and 100\% of the originally related reference text; and complementarily include 100\%, 75\%, 50\%, 25\%, 0\% of unrelated reference text respectively.

\begin{table*}[t]
\fontsize{8}{8}\selectfont
\centering
\begin{tabular}{l|ccccccc}
\multicolumn{1}{c|}{\textbf{Method}} & \textbf{ROUGE-1} & \textbf{ROUGE-2} & \textbf{ROUGE-L} & \textbf{METEOR} & \textbf{BLEU} & \textbf{BertScore} & \textbf{PARENT} \\ \hline
 \multicolumn{8}{|c|}{\textbf{Tested on WebNLG}} \\ \hline
Fully-supervised fine-tuning    & 59.99\textsubscript{(0.10)} & \textbf{40.93}\textsubscript{(0.18)} & \textbf{49.32}\textsubscript{(0.15)} & \textbf{39.76}\textsubscript{(0.04)} & \textbf{42.83}\textsubscript{(0.21)} & \textbf{95.41}\textsubscript{(0.02)} & 45.67\textsubscript{(0.30)} \\ \hline
Low-resource fine-tuning        & 55.55\textsubscript{(0.67)} & 36.63\textsubscript{(0.37)} & 46.21\textsubscript{(0.35)} & 35.22\textsubscript{(0.70)} & 33.63\textsubscript{(0.87)} & 94.60\textsubscript{(0.08)} & 41.37\textsubscript{(0.54)} \\ %
+ additional pre-training       & 55.28\textsubscript{(0.43)} & 35.71\textsubscript{(0.32)} & 45.41\textsubscript{(0.24)} & 35.26\textsubscript{(0.46)} & 33.44\textsubscript{(0.59)} & 94.33\textsubscript{(0.06)} & 39.47\textsubscript{(0.52)} \\ %
Unsupervised cycle training     & 58.65\textsubscript{(0.53)} & 37.70\textsubscript{(1.02)} & 46.18\textsubscript{(0.59)} & 37.98\textsubscript{(0.33)} & 36.36\textsubscript{(2.35)} & 94.42\textsubscript{(0.26)} & 43.24\textsubscript{(1.10)} \\ %
Low-resource cycle training        & \textbf{\underline{60.21}}\textsubscript{(0.21)} & \underline{40.56}\textsubscript{(0.42)} & \underline{48.71}\textsubscript{(0.17)} & \underline{39.74}\textsubscript{(0.32)} & \underline{41.77}\textsubscript{(0.70)} & \underline{95.18}\textsubscript{(0.04)} & \textbf{\underline{46.14}}\textsubscript{(0.36)} \\ \hline

 \multicolumn{8}{|c|}{\textbf{Tested on E2E}} \\ \hline
Fully-supervised fine-tuning    & \textbf{69.77}\textsubscript{(0.10)} & \textbf{42.87}\textsubscript{(0.17)} & \textbf{50.93}\textsubscript{(0.18)} & 52.90\textsubscript{(0.43)} & \textbf{29.35}\textsubscript{(0.47)} & \textbf{94.76}\textsubscript{(0.02)} & \textbf{41.91}\textsubscript{(0.61)} \\ \hline
Low-resource fine-tuning        & 66.62\textsubscript{(0.15)} & 39.68\textsubscript{(0.25)} & 48.59\textsubscript{(0.18)} & 48.80\textsubscript{(0.39)} & 25.31\textsubscript{(0.31)} & 94.35\textsubscript{(0.02)} & 39.56\textsubscript{(1.21)} \\ %
+ additional pre-training       & 66.88\textsubscript{(0.40)} & 39.45\textsubscript{(0.33)} & 48.65\textsubscript{(0.36)} & 50.11\textsubscript{(0.65)} & 26.29\textsubscript{(0.55)} & 94.35\textsubscript{(0.04)} & 39.65\textsubscript{(0.53)} \\ %
Unsupervised cycle training     & 63.43\textsubscript{(0.81)} & 37.73\textsubscript{(0.32)} & 45.96\textsubscript{(0.61)} & 50.49\textsubscript{(0.78)} & 27.92\textsubscript{(0.37)} & 93.71\textsubscript{(0.09)} & 37.97\textsubscript{(0.30)} \\ %
Low-resource cycle training        & \underline{69.53}\textsubscript{(0.25)} & \underline{42.48}\textsubscript{(0.20)} & \underline{50.51}\textsubscript{(0.28)} & \textbf{\underline{53.02}}\textsubscript{(0.24)} & \underline{29.22}\textsubscript{(0.12)} & \underline{94.74}\textsubscript{(0.02)} & \underline{41.39}\textsubscript{(0.70)} \\ \hline

 \multicolumn{8}{|c|}{\textbf{Tested on WTQ}} \\ \hline
Fully-supervised fine-tuning    & \textbf{62.25}\textsubscript{(0.66)} & \textbf{34.59}\textsubscript{(0.61)} & \textbf{49.41}\textsubscript{(0.57)} & \textbf{39.17}\textsubscript{(0.86)} & \textbf{21.18}\textsubscript{(0.53)} & \textbf{92.88}\textsubscript{(0.05)} & 24.18\textsubscript{(0.74)} \\ \hline
Low-resource fine-tuning        & 55.89\textsubscript{(0.88)} & 31.60\textsubscript{(0.81)} & 46.73\textsubscript{(0.64)} & 31.98\textsubscript{(0.57)} & 15.34\textsubscript{(0.72)} & 91.91\textsubscript{(0.14)} & 23.36\textsubscript{(1.05)} \\ %
+ additional pre-training       & 55.57\textsubscript{(0.68)} & 30.48\textsubscript{(0.80)} & 44.47\textsubscript{(0.74)} & 33.73\textsubscript{(0.74)} & 15.89\textsubscript{(0.39)} & 91.53\textsubscript{(0.17)} & 22.88\textsubscript{(0.43)} \\ %
Unsupervised cycle training     & 61.27\textsubscript{(0.50)} & 33.45\textsubscript{(0.52)} & 48.22\textsubscript{(0.44)} & 39.06\textsubscript{(0.22)} & 20.46\textsubscript{(0.69)} & \underline{92.67}\textsubscript{(0.04)} & 23.05\textsubscript{(0.35)} \\ %
Low-resource cycle training        & \underline{61.54}\textsubscript{(0.29)} & \underline{34.25}\textsubscript{(0.78)} & \underline{49.07}\textsubscript{(0.45)} & \underline{39.09}\textsubscript{(0.60)} & \underline{20.93}\textsubscript{(0.98)} & 92.66\textsubscript{(0.10)} & \textbf{\underline{24.39}}\textsubscript{(0.84)} \\ \hline

 \multicolumn{8}{|c|}{\textbf{Tested on WSQL}} \\ \hline
Fully-supervised fine-tuning    & 58.27\textsubscript{(1.79)} & 32.77\textsubscript{(1.15)} & 48.40\textsubscript{(2.44)} & \textbf{37.95}\textsubscript{(0.99)} & 22.97\textsubscript{(1.38)} & \textbf{93.18}\textsubscript{(0.19)} & 24.00\textsubscript{(2.07)} \\ \hline
Low-resource fine-tuning        & 56.37\textsubscript{(1.15)} & 31.60\textsubscript{(0.59)} & 49.42\textsubscript{(0.77)} & 33.57\textsubscript{(0.24)} & 23.34\textsubscript{(1.03)} & 92.57\textsubscript{(0.18)} & 23.68\textsubscript{(1.11)} \\ %
+ additional pre-training       & 56.01\textsubscript{(0.66)} & 30.92\textsubscript{(0.92)} & 47.00\textsubscript{(1.18)} & 35.34\textsubscript{(0.86)} & 21.18\textsubscript{(0.65)} & 92.24\textsubscript{(0.33)} & 22.66\textsubscript{(0.56)} \\ %
Unsupervised cycle training     & 42.24\textsubscript{(0.23)} & 15.17\textsubscript{(0.13)} & 33.52\textsubscript{(0.23)} & 29.45\textsubscript{(0.29)} & 4.03\textsubscript{(0.15)} & 85.37\textsubscript{(0.14)} & 14.63\textsubscript{(0.17)} \\ %
Low-resource cycle training        & \textbf{\underline{58.71}}\textsubscript{(1.43)} & \textbf{\underline{33.13}}\textsubscript{(1.90)} & \textbf{\underline{51.01}}\textsubscript{(1.43)} & \underline{37.43}\textsubscript{(1.04)} & \textbf{\underline{25.60}}\textsubscript{(1.58)} & \underline{93.03}\textsubscript{(0.18)} & \textbf{\underline{25.84}}\textsubscript{(1.42)} \\ \hline

\end{tabular}
\caption{Evaluation of data-to-text generation (\textbf{bold}: best of all; \underline{underlined}: best of low-resource settings). We report the average and standard deviation (in parenthesized subscripts) of each metric for 5 repeated runs.}
\label{table1}
\end{table*}

\subsection{Training Parameters}
\label{subsec:parameters}
We use the T5-base model which has 12 layers, a hidden size of 768, 12 self-attention heads, and 220M parameters. We use the AdamW optimizer with linear weight decay, a max input length of 256, a learning rate of 3e-4, and an effective batch size of 256. At inference time, we decode with the beam search algorithm using 4 beams and a generation length varying between 3 tokens and 256 tokens. 
We train each model up to 50 epochs with a delta of 0.05 basis points and a patience of 5 epochs as the early stopping criteria. We select the best model by the validation set's METEOR score - the ranking metric of the WebNLG 2020 challenge, and we report the aforementioned model's performance on the test set. We repeat each experiment 5 times with different random seeds and report the average and standard deviation of each metric.

\section{Results and Discussion}
\label{sec:results-n-discussion}

\begin{table*}[t]
\fontsize{8}{8}\selectfont
\centering
\begin{tabular}{c|ccccccc}
\multicolumn{1}{c|}{\textbf{Overlapping Level}} & \textbf{ROUGE-1} & \textbf{ROUGE-2} & \textbf{ROUGE-L} & \textbf{METEOR} & \textbf{BLEU} & \textbf{BertScore} & \textbf{PARENT} \\ \hline
0\%    & 52.50\textsubscript{(0.43)} & 31.16\textsubscript{(0.40)} & 40.14\textsubscript{(0.46)} & 35.99\textsubscript{(0.46)} & 26.69\textsubscript{(1.03)} & 92.59\textsubscript{(0.12)} & 34.33\textsubscript{(0.58)} \\ %
25\%        & 56.23\textsubscript{(0.67)} & 34.59\textsubscript{(0.82)} & 43.46\textsubscript{(0.63)} & 37.23\textsubscript{(0.17)} & 32.21\textsubscript{(1.74)} & 93.63\textsubscript{(0.22)} & 39.28\textsubscript{(0.96)} \\ %
50\%       & 58.64\textsubscript{(0.34)} & 37.40\textsubscript{(0.60)} & 46.05\textsubscript{(0.41)} & 38.07\textsubscript{(0.28)} & 35.83\textsubscript{(1.07)} & 94.41\textsubscript{(0.17)} & 43.09\textsubscript{(0.68)} \\ %
75\%     & 58.64\textsubscript{(0.32)} & 37.66\textsubscript{(0.26)} & 46.36\textsubscript{(0.23)} & 37.78\textsubscript{(0.18)} & 36.91\textsubscript{(0.37)} & 94.46\textsubscript{(0.09)} & 43.47\textsubscript{(0.37)} \\ %
100\%        & 58.75\textsubscript{(0.28)} & 38.04\textsubscript{(0.44)} & 46.44\textsubscript{(0.19)} & 37.86\textsubscript{(0.25)} & 37.39\textsubscript{(0.79)} & 94.57\textsubscript{(0.12)} & 43.76\textsubscript{(0.32)} \\ \hline
\end{tabular}
\caption{Cycle training with the same amount of data at different overlapping levels. We report the average and standard deviation (in parenthesized subscripts) of each metric for 5 repeated runs.}
\label{table22}
\end{table*}

\subsection{Automatic Evaluation}
\label{subsec:automatic-eval}
We assess each system/strategy with five widely-used automatic metrics that measure the generation quality from three different aspects: token-matching, semantic similarity, and faithfulness. 

\noindent\textbf{ROUGE} \cite{lin-2004-rouge} is a recall-oriented metric that calculates the overlapping n-grams (ROUGE-N for N-grams) and word sequences (ROUGE-L) between the reference text and generated text.

\noindent\textbf{BLEU} \cite{papineni-etal-2002-bleu} is a precision-oriented metric calculating overlapping n-grams between the reference text and generated text. 

\noindent\textbf{METEOR} \cite{banerjee-lavie-2005-meteor} computes the unigram match between the reference text and generated text based on the tokens' surface form, stemming, synonyms, and paraphrase similarities. 

\noindent\textbf{BertScore} \cite{bert-score} measures the semantic similarity of the reference text and generated text via the utilization of the contextual embeddings from BERT for the calculation of the cosine similarity of best-matching token pairs.

\noindent\textbf{PARENT} \cite{PARENT-metric} is an entailment-based token-matching metric that calculates the F1 score based on entailed precision (an n-gram is correct if it occurs in the reference text or entailed by the input data) and entailed recall (recall against the reference text input data, adjusted by a weight parameter). It measures the faithfulness of the generated text with respect to the input data. 

Table \ref{table1} displays the performance of multiple data-to-text generation approaches under various settings. We observe that unsupervised cycle training generally falls short of the fully-supervised fine-tuning method's performance. When compared with the low-resource fine-tuning method, it scored higher on WebNLG and WTQ but performed worse on E2E and WSQL, where the performance gap on WSQL is larger. We attribute such divergence to the difference in the number of unique predicates and vocabulary. Cycle training should be able to improve the model's generalizability and robustness through exposure to larger amounts of diverse text and structured data, and through its capability of gradually learning different data-to-text associations. For datasets like E2E and WSQL, their smaller vocabulary size and number of unique predicates imply that a small amount of annotated samples might cover a great deal of the datasets' underlying variation. This leads to a strong low-resource fine-tuning performance that has smaller performance gaps with the fully-supervised counterparts, and overshadows the unsupervised cycle training method. 

However, when a small amount of annotated data is made available for initializing the cycle training, the low-resource cycle training strategy significantly improves the generation performance over the low-resource fine-tuning method, and achieves competitive performance with respect to the fully-supervised method. Such an improvement is consistent across all four datasets and five types of evaluation metrics. 
Notably, when applied to multi-domain and open-domain datasets (WebNLG, WTQ, and WSQL), low-resource cycle training generated texts that have better faithfulness to the input data, evident from the PARENT score, compared to the fully-supervised fine-tuning approach. Compared with the setting that applies additional pre-training, it is evident that cycle training works beyond simply raising the probability of generating target domain tokens.

As for the experiments on cycle training with unpaired datasets at different overlapping levels, the results in Table \ref{table22} show that performance sharply increases at the beginning with the increase of overlapping levels and then turns to flatten at around the 50\% overlapping level. This suggests that when the size is the same, the unpaired data corpus and text corpus used for cycle training need to have at least 50\% entities (or say, latent information) overlap to achieve performance at an ideal level. We deem 50\% as a reasonable level since many related but unpaired texts and structured data (e.g., content and infoboxes from Wikipedia, product specification tables and descriptions from online shopping platforms, etc.) may have higher information overlap. 
Hence, based on our experimental results, we believe that low-resource cycle training is a universally applicable approach that can effectively learn from vast unpaired structured data and texts with minimal human effort.

\subsection{Human Evaluation}
\label{subsec:human-eval}

\begin{table}[t]
\footnotesize
\centering
\begin{tabular}{|ccccc|}
\hline
\multicolumn{1}{|c|}{\textbf{Method}} &
  \multicolumn{1}{c|}{\textbf{FE}} &
  \multicolumn{1}{c|}{\textbf{HE}} &
  \multicolumn{1}{c|}{\textbf{IM}} &
  \textbf{FP} \\ \hline \hline
\multicolumn{5}{|c|}{\textbf{Combined}} \\ \hline
\multicolumn{1}{|c|}{\begin{tabular}[c]{@{}c@{}}Low-resource\\ fine-tuning\end{tabular}} &
  \multicolumn{1}{c|}{8.05} &
  \multicolumn{1}{c|}{14.84} &
  \multicolumn{1}{c|}{21.39} &
  {2.00} \\ \hline
\multicolumn{1}{|c|}{\begin{tabular}[c]{@{}c@{}}Low-resource\\ cycle-training\end{tabular}} &
  \multicolumn{1}{c|}{\textbf{0.49}} &
  \multicolumn{1}{c|}{\textbf{2.57}} &
  \multicolumn{1}{c|}{\textbf{3.36}} &
  {1.80} \\ \hline
\multicolumn{1}{|c|}{\begin{tabular}[c]{@{}c@{}}Fully-supervised\\ fine-tuning\end{tabular}} &
  \multicolumn{1}{c|}{2.08} &
  \multicolumn{1}{c|}{11.48} &
  \multicolumn{1}{c|}{8.46} &
  {\textbf{1.73}} \\ \hline

\multicolumn{5}{|c|}{\textbf{WebNLG}} \\ \hline
\multicolumn{1}{|c|}{\begin{tabular}[c]{@{}c@{}}Low-resource\\ fine-tuning\end{tabular}} &
  \multicolumn{1}{c|}{6.72} &
  \multicolumn{1}{c|}{7.21} &
  \multicolumn{1}{c|}{15.90} &
  {1.91} \\ \hline
\multicolumn{1}{|c|}{\begin{tabular}[c]{@{}c@{}}Low-resource\\ cycle-training\end{tabular}} &
  \multicolumn{1}{c|}{\textbf{0.00}} &
  \multicolumn{1}{c|}{\textbf{1.47}} &
  \multicolumn{1}{c|}{\textbf{1.82}} &
  {1.89} \\ \hline
\multicolumn{1}{|c|}{\begin{tabular}[c]{@{}c@{}}Fully-supervised\\ fine-tuning\end{tabular}} &
  \multicolumn{1}{c|}{\textbf{0.00}} &
  \multicolumn{1}{c|}{6.72} &
  \multicolumn{1}{c|}{10.29} &
  {\textbf{1.73}} \\ \hline

\multicolumn{5}{|c|}{\textbf{E2E}} \\ \hline
\multicolumn{1}{|c|}{\begin{tabular}[c]{@{}c@{}}Low-resource\\ fine-tuning\end{tabular}} &
  \multicolumn{1}{c|}{\textbf{0.00}} &
  \multicolumn{1}{c|}{1.18} &
  \multicolumn{1}{c|}{6.43} &
  {1.99} \\ \hline
\multicolumn{1}{|c|}{\begin{tabular}[c]{@{}c@{}}Low-resource\\ cycle-training\end{tabular}} &
  \multicolumn{1}{c|}{\textbf{0.00}} &
  \multicolumn{1}{c|}{\textbf{0.00}} &
  \multicolumn{1}{c|}{0.84} &
  {1.86} \\ \hline
\multicolumn{1}{|c|}{\begin{tabular}[c]{@{}c@{}}Fully-supervised\\ fine-tuning\end{tabular}} &
  \multicolumn{1}{c|}{\textbf{0.00}} &
  \multicolumn{1}{c|}{\textbf{0.00}} &
  \multicolumn{1}{c|}{\textbf{0.00}} &
  {\textbf{1.64}} \\ \hline

\multicolumn{5}{|c|}{\textbf{WTQ}} \\ \hline
\multicolumn{1}{|c|}{\begin{tabular}[c]{@{}c@{}}Low-resource\\ fine-tuning\end{tabular}} &
  \multicolumn{1}{c|}{14.71} &
  \multicolumn{1}{c|}{15.69} &
  \multicolumn{1}{c|}{33.82} &
  {2.16} \\ \hline
\multicolumn{1}{|c|}{\begin{tabular}[c]{@{}c@{}}Low-resource\\ cycle-training\end{tabular}} &
  \multicolumn{1}{c|}{\textbf{0.00}} &
  \multicolumn{1}{c|}{\textbf{0.00}} &
  \multicolumn{1}{c|}{\textbf{1.96}} &
  {\textbf{1.75}} \\ \hline
\multicolumn{1}{|c|}{\begin{tabular}[c]{@{}c@{}}Fully-supervised\\ fine-tuning\end{tabular}} &
  \multicolumn{1}{c|}{8.33} &
  \multicolumn{1}{c|}{24.51} &
  \multicolumn{1}{c|}{8.82} &
  {1.85} \\ \hline

\multicolumn{5}{|c|}{\textbf{WSQL}} \\ \hline
\multicolumn{1}{|c|}{\begin{tabular}[c]{@{}c@{}}Low-resource\\ fine-tuning\end{tabular}} &
  \multicolumn{1}{c|}{10.78} &
  \multicolumn{1}{c|}{35.29} &
  \multicolumn{1}{c|}{29.41} &
  {1.93} \\ \hline
\multicolumn{1}{|c|}{\begin{tabular}[c]{@{}c@{}}Low-resource\\ cycle-training\end{tabular}} &
  \multicolumn{1}{c|}{1.96} &
  \multicolumn{1}{c|}{\textbf{8.82}} &
  \multicolumn{1}{c|}{\textbf{8.82}} &
  {\textbf{1.72}} \\ \hline
\multicolumn{1}{|c|}{\begin{tabular}[c]{@{}c@{}}Fully-supervised\\ fine-tuning\end{tabular}} &
  \multicolumn{1}{c|}{\textbf{0.00}} &
  \multicolumn{1}{c|}{14.71} &
  \multicolumn{1}{c|}{14.71} &
  {1.76} \\ \hline

\end{tabular}
\caption{Normalized Human Evaluation Results (lower is better; \textbf{bold}: best of all.)}
\label{table4}
\end{table}

\begin{table*}[t]
\scriptsize
\centering
\begin{tabular}{|c|l|c|} 
\hline
\textbf{Input Triple Set}                                             & \multicolumn{2}{l|}{\begin{tabular}[c]{@{}l@{}}1. \textcolor[rgb]{0.259,0.522,0.957}{[S] Liselotte Grschebina [P] birthplace [O] Karlsruhe}\\2. \textcolor[rgb]{0.918,0.263,0.208}{[S] Liselotte Grschebina [P] nationality [O] Israel}\\3. \textcolor[rgb]{0.984,0.737,0.016}{[S] Liselotte Grschebina [P] training [O] Schoolof Applied Arts in Stuttgart}\\4. \textcolor[rgb]{0.204,0.659,0.325}{[S] Karlsruhe [P] country [O] Germany}\\5. \textcolor[rgb]{1,0.427,0.004}{[S] Israel [P] language [O] Modern Hebrew}\end{tabular}}                                                                                                                             \\ 
\hline
\textbf{Reference Text}                                               & \multicolumn{2}{l|}{\begin{tabular}[c]{@{}l@{}}- Liselotte Grschebina was an Israeli who attended the school of Applied Arts in Stuttgart. She was born \\in Karlsruhe, Germany. Israel's national language is Modern Hebrew.\\- Liselotte Grschebina is an Israeli national who did her studies at the School of Applied Arts in Stuttgart \\and was born in Karlsruhe, which is in Germany unlike Israel whose language is Modern Hebrew.\\- Liselotte Grschebina was born in Karlsruhe which is located in Germany, has Israeli nationality (the \\language in Israel is Modern Hebrew), and studied at the School of Applied Arts in Stuttgart.\end{tabular}}  \\ 
\hline
\textbf{Method}                                                       & \multicolumn{1}{c|}{\textbf{Generation}}                                                                                                                                                                                                                                                                                                                                                        & \textbf{Remark}                                                                                                                                                                                                                                              \\ 
\hline
\begin{tabular}[c]{@{}c@{}}Fully-supervised\\fine-tuning\end{tabular} & \begin{tabular}[c]{@{}l@{}}\textcolor[rgb]{0.259,0.522,0.957}{Liselotte Grschebina was born in Karlsruhe}\textcolor[rgb]{0.204,0.659,0.325}{, Germany.} \textcolor[rgb]{0.984,0.737,0.016}{She studied at the School of Applied }\\\textcolor[rgb]{0.984,0.737,0.016}{Arts in Stuttgart.} \textcolor[rgb]{1,0.427,0.004}{Modern Hebrew is spoken in Israel.}\end{tabular}                       & 1 IM                                                                                                                                                                                                                                                             \\ 
\hline
\begin{tabular}[c]{@{}c@{}}Low-resource\\fine-tuning\end{tabular}     & \begin{tabular}[c]{@{}l@{}}\textcolor[rgb]{0.259,0.522,0.957}{Liselotte Grschebina was born in Karlsruhe}\textcolor[rgb]{0.204,0.659,0.325}{, }\textcolor[rgb]{0.204,0.659,0.325}{Israel} \textcolor[rgb]{0.984,0.737,0.016}{and trained at the School of Applied Arts }\\\textcolor[rgb]{0.984,0.737,0.016}{in Stuttgart.}\end{tabular}                                                        & 3 IMs, 1 FE                                                                                                                                                                                                                                                      \\ 
\hline
\begin{tabular}[c]{@{}c@{}}Low-resource\\cycle-training\end{tabular}  & \begin{tabular}[c]{@{}l@{}}\textcolor[rgb]{0.259,0.522,0.957}{Liselotte Grschebina was born in Karlsruhe}\textcolor[rgb]{0.204,0.659,0.325}{, Germany} \textcolor[rgb]{0.984,0.737,0.016}{and was a student at the School of }\\\textcolor[rgb]{0.984,0.737,0.016}{Applied Arts in Stuttgart.} \textcolor[rgb]{1,0.427,0.004}{Modern Hebrew is spoken in Israel.}\end{tabular}                  & 1 IM                                                                                                                                                                                                                                                             \\ 
\hline
\end{tabular}
\caption{Error analysis example.}
\label{table6}
\end{table*}

To quantitatively compare generated text with respect to correctness, faithfulness, data coverage, and fluency, we develop a new counting and ranking-based annotation schema, and use it to conduct human evaluation. Our schema features better objectiveness, consistency, and precision compared to the 0-100 rating-based schema used for the WebNLG 2020 Challenge. We define the following measures (full annotation guidelines, including disambiguation examples, and screenshots of the annotation interface available in Appendix~\ref{sec:guidelines}):

\smallskip

\noindent\textbf{Count of Factual Errors (FE)} measures the factual correctness of the generated text with respect to the entities (subject and object) and predicates of the input triplets. Factual errors are information in the generations that contradict the information in the input subject-predicate-object context. For each attempted predicate given in the input triplets, the annotator is asked to increase the factual error count if the subject and/or object of the predicate’s associated expression doesn't match facts from the input. 

\noindent\textbf{Count of Hallucination Errors (HE)} measures the relevance of the generated text with respect to the input triplets. Hallucination errors occur when words or phrases in the generation cannot be inferred from the input subject-predicate-object triplets, for instance, because the value does not make logical sense, or because the predicate of the expression is not present in any triple. Unlike FEs, HEs add information  not present in the triplets or reference, but do not directly contradict the triplets. The annotator is asked to increase the HE count if a piece of information contained in the generated text is not presented in, or cannot be \textit{reasonably inferred} by the input triplets. For better consistency and less ambiguity, a \textit{reasonable inference} is defined as a piece of information contained in the generated text that isn't present in the input triplets but is present in the reference text.

\noindent\textbf{Count of Information Misses (IM)} measures the information coverage of the generated text with respect to the predicates given in the input triplets. For each predicate given in the input triplets, the annotator is asked to increase the IM count if the generated text does not attempt to express the predicate. 

\noindent\textbf{Fluency Preference (FP)} measures the quality of the generated text in terms of the grammar, structure, and coherence of the text. The annotator is asked to compare the fluency of pairs of generated texts within a batch, to compile the final ranking that reflects the annotator's subjective preference. The fluency comparison and ranking only considers the grammar, structure, and coherence of the text independent of IM, FE, and HE.

In terms of the training time required to perform the task accurately, we collected the error annotations (FE, HE, IM) from two domain experts and the fluency annotations from crowd-sourced workers respectively via an annotation tool built on the Appen\footnote{\url{https://appen.com/}} platform. To enforce the annotation quality and foster future research on explainable automatic error analysis, we ask the domain experts to mark the token(s) that constitute an FE or HE, and to select the triple(s) that constitute the IM before counting the respective errors. The domain experts independently annotate the same set of 204 randomly sampled generations with a resulting agreement (Cohen’s kappa score \cite{artstein-poesio-2008-survey}) of 0.74 for FE, 0.69 for HE, and 0.85 for IM, which is very satisfactory given the complexity of the task. For the relatively more subjective fluency ranking task, we use the average of three crowd-sourced native English speakers' judgments for each generation. As generating longer text for larger triple sets is more difficult than generating for smaller triplets, we normalize the counts of FE, HE, and IM by the number of their input triples. Therefore, the FE, HE, and IM we report in Table~\ref{table4} can be interpreted as the probability of making such errors per input data triple. 
We show an example of our error analysis in Table \ref{table6}, and provide additional examples in Appendix~\ref{sec:appendix}. 

Our human evaluation suggests that low-resource cycle training consistently reduces factual errors, hallucination errors and information misses. From Section \ref{subsec:automatic-eval}, cycle training presents a larger performance gain when applied to datasets that have more variations in terms of underlying relations and surface realizations. When looking together with Table \ref{table1}, the human evaluation of errors and information coverage correlates better with the PARENT score, which confirms PARENT's capability of measuring faithfulness. It is also evident from the annotation results that all three evaluated data-to-text generation models are more likely to make hallucination errors over factual errors, which calls for more future effort to alleviate hallucinations. In terms of the generated texts' fluency, low-resource cycle training is able to improve over the low-resource fine-tuning method but still cannot consistently beat the fully-supervised approach.

\section{Conclusions}
\label{sec:conclusions}

In this work, we demonstrated the application of cycle training for data-to-text generation. 
We systematically investigated the effectiveness of cycle training across different domains, and the application of pre-cycle fine-tuning in low-resource settings. We showed that our approach substantially improved data-to-text generation performance in low-resource settings, achieved competitive performance compared to fully-supervised models, and also improved the faithfulness of the generated text through a reduction in factual errors, hallucinations and information misses, even when compared to fully supervised approaches. 
We also designed a schema for effective human evaluation of data-to-text generation, that improves upon prior work and encourages more objective and consistent reviews of faithfulness. 

\section*{Limitations}

We recognize that our annotation and analysis methods can require considerable human labor, that can limit the amount of annotated data we can collect. 
Also, despite cycle training being generally accepted as a model-agnostic approach, we were not able to test a wide variety of backbone models due to resource constraints.
In addition, though we relaxed the entity constraints and made cycle training for data-to-text generation end-to-end, the non-differentiability problem remains unsolved. The intermediate outputs generated by the first model of each cycle are assumed to be correct. This is a weak assumption that may propagate misleading training signals to the second model of each cycle, particularly in the early stage of the training.

To address these limitations, future work may focus on the following directions: 1) building differentiable cycle training models; 2) exploring automated error detection methods and building models that may utilize such signals; and 3) assessing different backbone models, including large language models like GPT-X, with the cycle training approach.

\section*{Acknowledgements}
First and foremost, we extend our appreciation to Prof. James Caverlee for his unwavering support that was vital for the completion of this work. We gratefully acknowledge the contributions of the following individuals for their expert advice as well as their participation in our preliminary human annotation study, which helped us a lot in refining our experiments, annotation guidelines and annotation interface: Dr. Giuseppe Castellucci, Dr. Besnik Fetahu, Prof. Eugene Agichtein, Dr. Saar Kuzi, Jason Ingyu Choi, Dr. Zhiyu Chen, Dr. Tuan M. Lai, Lingbo Mo, and Yicheng Wang. We also would like to express our gratitude to the three reviewers and the meta reviewer for their constructive suggestions.

\bibliography{project}

\begin{thebibliography}{41}
\expandafter\ifx\csname natexlab\endcsname\relax\def\natexlab#1{#1}\fi

\bibitem[{Agarwal et~al.(2020)Agarwal, Kale, Ge, Shakeri, and
  Al-Rfou}]{agarwal-etal-2020-machine}
Oshin Agarwal, Mihir Kale, Heming Ge, Siamak Shakeri, and Rami Al-Rfou. 2020.
\newblock \href {https://aclanthology.org/2020.webnlg-1.13} {Machine
  translation aided bilingual data-to-text generation and semantic parsing}.
\newblock In \emph{Proceedings of the 3rd International Workshop on Natural
  Language Generation from the Semantic Web (WebNLG+)}, pages 125--130, Dublin,
  Ireland (Virtual). Association for Computational Linguistics.

\bibitem[{Artstein and Poesio(2008)}]{artstein-poesio-2008-survey}
Ron Artstein and Massimo Poesio. 2008.
\newblock \href {https://doi.org/10.1162/coli.07-034-R2} {Survey article:
  Inter-coder agreement for computational linguistics}.
\newblock \emph{Computational Linguistics}, 34(4):555--596.

\bibitem[{Banerjee and Lavie(2005)}]{banerjee-lavie-2005-meteor}
Satanjeev Banerjee and Alon Lavie. 2005.
\newblock \href {https://aclanthology.org/W05-0909} {{METEOR}: An automatic
  metric for {MT} evaluation with improved correlation with human judgments}.
\newblock In \emph{Proceedings of the {ACL} Workshop on Intrinsic and Extrinsic
  Evaluation Measures for Machine Translation and/or Summarization}, pages
  65--72, Ann Arbor, Michigan. Association for Computational Linguistics.

\bibitem[{Brown et~al.(2020)Brown, Mann, Ryder, Subbiah, Kaplan, Dhariwal,
  Neelakantan, Shyam, Sastry, Askell, Agarwal, Herbert-Voss, Krueger, Henighan,
  Child, Ramesh, Ziegler, Wu, Winter, Hesse, Chen, Sigler, Litwin, Gray, Chess,
  Clark, Berner, McCandlish, Radford, Sutskever, and
  Amodei}]{NEURIPS2020_1457c0d6}
Tom Brown, Benjamin Mann, Nick Ryder, Melanie Subbiah, Jared~D Kaplan, Prafulla
  Dhariwal, Arvind Neelakantan, Pranav Shyam, Girish Sastry, Amanda Askell,
  Sandhini Agarwal, Ariel Herbert-Voss, Gretchen Krueger, Tom Henighan, Rewon
  Child, Aditya Ramesh, Daniel Ziegler, Jeffrey Wu, Clemens Winter, Chris
  Hesse, Mark Chen, Eric Sigler, Mateusz Litwin, Scott Gray, Benjamin Chess,
  Jack Clark, Christopher Berner, Sam McCandlish, Alec Radford, Ilya Sutskever,
  and Dario Amodei. 2020.
\newblock \href
  {https://proceedings.neurips.cc/paper_files/paper/2020/file/1457c0d6bfcb4967418bfb8ac142f64a-Paper.pdf}
  {Language models are few-shot learners}.
\newblock In \emph{Advances in Neural Information Processing Systems},
  volume~33, pages 1877--1901. Curran Associates, Inc.

\bibitem[{Castro~Ferreira et~al.(2020)Castro~Ferreira, Gardent, Ilinykh,
  van~der Lee, Mille, Moussallem, and Shimorina}]{WebNLG2020report}
Thiago Castro~Ferreira, Claire Gardent, Nikolai Ilinykh, Chris van~der Lee,
  Simon Mille, Diego Moussallem, and Anastasia Shimorina. 2020.
\newblock \href {https://aclanthology.org/2020.webnlg-1.7} {The 2020 bilingual,
  bi-directional {W}eb{NLG}+ shared task: Overview and evaluation results
  ({W}eb{NLG}+ 2020)}.
\newblock In \emph{Proceedings of the 3rd International Workshop on Natural
  Language Generation from the Semantic Web (WebNLG+)}, pages 55--76, Dublin,
  Ireland (Virtual). Association for Computational Linguistics.

\bibitem[{Chen et~al.(2021)Chen, Wiseman, and Gimpel}]{WikiTableT}
Mingda Chen, Sam Wiseman, and Kevin Gimpel. 2021.
\newblock \href {https://doi.org/10.18653/v1/2021.findings-acl.17}
  {{WikiTableT: A Large-Scale Data-to-Text Dataset for Generating Wikipedia
  Article Sections}}.
\newblock \emph{Findings of the Association for Computational Linguistics:
  ACL-IJCNLP 2021}, pages 193--209.

\bibitem[{Chen et~al.(2019)Chen, Wang, Chen, Zhang, Wang, Li, Zhou, and
  Wang}]{2019TabFactA}
Wenhu Chen, Hongmin Wang, Jianshu Chen, Yunkai Zhang, Hong Wang, Shiyang Li,
  Xiyou Zhou, and William~Yang Wang. 2019.
\newblock {TabFact: A Large-scale Dataset for Table-based Fact Verification}.
\newblock In \emph{International Conference on Learning Representations
  (ICLR)}, arXiv, Addis Ababa, Ethiopia.

\bibitem[{Colin et~al.(2016)Colin, Gardent, Mrabet, Narayan, and
  Perez-Beltrachini}]{WebNLG-DBPedia}
Emilie Colin, Claire Gardent, Yassine Mrabet, Shashi Narayan, and Laura
  Perez-Beltrachini. 2016.
\newblock \href {https://doi.org/10.18653/v1/w16-6626} {{The WebNLG Challenge:
  Generating Text from DBPedia Data}}.
\newblock \emph{Proceedings of the 9th International Natural Language
  Generation conference}, pages 163--167.

\bibitem[{Devlin et~al.(2019)Devlin, Chang, Lee, and
  Toutanova}]{devlin-etal-2019-bert}
Jacob Devlin, Ming-Wei Chang, Kenton Lee, and Kristina Toutanova. 2019.
\newblock \href {https://doi.org/10.18653/v1/N19-1423} {{BERT}: Pre-training of
  deep bidirectional transformers for language understanding}.
\newblock In \emph{Proceedings of the 2019 Conference of the North {A}merican
  Chapter of the Association for Computational Linguistics: Human Language
  Technologies, Volume 1 (Long and Short Papers)}, pages 4171--4186,
  Minneapolis, Minnesota. Association for Computational Linguistics.

\bibitem[{Dhingra et~al.(2019)Dhingra, Faruqui, Parikh, Chang, Das, and
  Cohen}]{PARENT-metric}
Bhuwan Dhingra, Manaal Faruqui, Ankur Parikh, Ming-Wei Chang, Dipanjan Das, and
  William~W Cohen. 2019.
\newblock \href {http://arxiv.org/abs/1906.01081} {{Handling Divergent
  Reference Texts when Evaluating Table-to-Text Generation}}.
\newblock \emph{arXiv}.
\newblock This is the PARENT evaluation metric paper.

\bibitem[{Estes et~al.(2022)Estes, Vedula, Collins, Cecil, and
  Rokhlenko}]{vedula2022emnlp}
Alex Estes, Nikhita Vedula, Marcus Collins, Matthew Cecil, and Oleg Rokhlenko.
  2022.
\newblock {Fact Checking Machine Generated Text with Dependency Trees}.
\newblock \emph{Proceedings of the 2022 Conference on Empirical Methods in
  Natural Language Processing (EMNLP)}.

\bibitem[{Gardent et~al.(2017{\natexlab{a}})Gardent, Shimorina, Narayan, and
  Perez-Beltrachini}]{WebNLG-creation}
Claire Gardent, Anastasia Shimorina, Shashi Narayan, and Laura
  Perez-Beltrachini. 2017{\natexlab{a}}.
\newblock \href {https://doi.org/10.18653/v1/p17-1017} {{Creating Training
  Corpora for NLG Micro-Planners}}.
\newblock \emph{Proceedings of the 55th Annual Meeting of the Association for
  Computational Linguistics (Volume 1: Long Papers)}, pages 179--188.

\bibitem[{Gardent et~al.(2017{\natexlab{b}})Gardent, Shimorina, Narayan, and
  Perez-Beltrachini}]{WebNLG-RDF}
Claire Gardent, Anastasia Shimorina, Shashi Narayan, and Laura
  Perez-Beltrachini. 2017{\natexlab{b}}.
\newblock \href {https://doi.org/10.18653/v1/w17-3518} {{The WebNLG Challenge:
  Generating Text from RDF Data}}.
\newblock \emph{Proceedings of the 10th International Conference on Natural
  Language Generation}, pages 124--133.

\bibitem[{Guo et~al.(2020)Guo, Jin, Qiu, Zhang, Wipf, and Zhang}]{CycleGT}
Qipeng Guo, Zhijing Jin, Xipeng Qiu, Weinan Zhang, David Wipf, and Zheng Zhang.
  2020.
\newblock \href {https://aclanthology.org/2020.webnlg-1.8} {{C}ycle{GT}:
  Unsupervised graph-to-text and text-to-graph generation via cycle training}.
\newblock In \emph{Proceedings of the 3rd International Workshop on Natural
  Language Generation from the Semantic Web (WebNLG+)}, pages 77--88, Dublin,
  Ireland (Virtual). Association for Computational Linguistics.

\bibitem[{Herzig et~al.(2020)Herzig, Nowak, Müller, Piccinno, and
  Eisenschlos}]{TAPAS}
Jonathan Herzig, Pawel~Krzysztof Nowak, Thomas Müller, Francesco Piccinno, and
  Julian Eisenschlos. 2020.
\newblock \href {https://doi.org/10.18653/v1/2020.acl-main.398} {{TaPas: Weakly
  Supervised Table Parsing via Pre-training}}.
\newblock \emph{Proceedings of the 58th Annual Meeting of the Association for
  Computational Linguistics}, pages 4320--4333.

\bibitem[{Hoang et~al.(2018)Hoang, Koehn, Haffari, and
  Cohn}]{hoang2018iterative}
Vu~Cong~Duy Hoang, Philipp Koehn, Gholamreza Haffari, and Trevor Cohn. 2018.
\newblock Iterative back-translation for neural machine translation.
\newblock In \emph{Proceedings of the 2nd workshop on neural machine
  translation and generation}, pages 18--24.

\bibitem[{Iovine et~al.(2022{\natexlab{a}})Iovine, Fang, Fetahu, Rokhlenko, and
  Malmasi}]{CycleNER}
Andrea Iovine, Anjie Fang, Besnik Fetahu, Oleg Rokhlenko, and Shervin Malmasi.
  2022{\natexlab{a}}.
\newblock \href {https://doi.org/10.1145/3485447.3512012} {{CycleNER: An
  Unsupervised Training Approach for Named Entity Recognition}}.
\newblock \emph{Proceedings of the ACM Web Conference 2022}, pages 2916--2924.

\bibitem[{Iovine et~al.(2022{\natexlab{b}})Iovine, Fang, Fetahu, Zhao,
  Rokhlenko, and Malmasi}]{iovine-etal-2022-cyclekqr}
Andrea Iovine, Anjie Fang, Besnik Fetahu, Jie Zhao, Oleg Rokhlenko, and Shervin
  Malmasi. 2022{\natexlab{b}}.
\newblock \href {https://aclanthology.org/2022.emnlp-main.814} {{C}ycle{KQR}:
  Unsupervised bidirectional keyword-question rewriting}.
\newblock In \emph{Proceedings of the 2022 Conference on Empirical Methods in
  Natural Language Processing}, pages 11875--11886, Abu Dhabi, United Arab
  Emirates. Association for Computational Linguistics.

\bibitem[{Lamb et~al.(2016)Lamb, ALIAS PARTH~GOYAL, Zhang, Zhang, Courville,
  and Bengio}]{lamb2016professor}
Alex~M Lamb, Anirudh~Goyal ALIAS PARTH~GOYAL, Ying Zhang, Saizheng Zhang,
  Aaron~C Courville, and Yoshua Bengio. 2016.
\newblock Professor forcing: A new algorithm for training recurrent networks.
\newblock \emph{Advances in neural information processing systems}, 29.

\bibitem[{Lample et~al.(2017)Lample, Conneau, Denoyer, and
  Ranzato}]{UnsuperNMT}
Guillaume Lample, Alexis Conneau, Ludovic Denoyer, and Marc'Aurelio Ranzato.
  2017.
\newblock {Unsupervised Machine Translation Using Monolingual Corpora Only}.
\newblock \emph{arXiv}.

\bibitem[{Lewis et~al.(2020)Lewis, Liu, Goyal, Ghazvininejad, Mohamed, Levy,
  Stoyanov, and Zettlemoyer}]{lewis-etal-2020-bart}
Mike Lewis, Yinhan Liu, Naman Goyal, Marjan Ghazvininejad, Abdelrahman Mohamed,
  Omer Levy, Veselin Stoyanov, and Luke Zettlemoyer. 2020.
\newblock \href {https://doi.org/10.18653/v1/2020.acl-main.703} {{BART}:
  Denoising sequence-to-sequence pre-training for natural language generation,
  translation, and comprehension}.
\newblock In \emph{Proceedings of the 58th Annual Meeting of the Association
  for Computational Linguistics}, pages 7871--7880, Online. Association for
  Computational Linguistics.

\bibitem[{Lin(2004)}]{lin-2004-rouge}
Chin-Yew Lin. 2004.
\newblock \href {https://aclanthology.org/W04-1013} {{ROUGE}: A package for
  automatic evaluation of summaries}.
\newblock In \emph{Text Summarization Branches Out}, pages 74--81, Barcelona,
  Spain. Association for Computational Linguistics.

\bibitem[{Liu et~al.(2021)Liu, Chen, Guo, Ziyadi, Lin, Chen, and Lou}]{TAPEX}
Qian Liu, Bei Chen, Jiaqi Guo, Morteza Ziyadi, Zeqi Lin, Weizhu Chen, and
  Jian-Guang Lou. 2021.
\newblock \href {http://arxiv.org/abs/2107.07653} {{TAPEX: Table Pre-training
  via Learning a Neural SQL Executor}}.
\newblock \emph{arXiv}.

\bibitem[{Malmasi et~al.(2022)Malmasi, Fang, Fetahu, Kar, and
  Rokhlenko}]{malmasi-etal-2022-semeval}
Shervin Malmasi, Anjie Fang, Besnik Fetahu, Sudipta Kar, and Oleg Rokhlenko.
  2022.
\newblock \href {https://doi.org/10.18653/v1/2022.semeval-1.196}
  {{S}em{E}val-2022 task 11: Multilingual complex named entity recognition
  ({M}ulti{C}o{NER})}.
\newblock In \emph{Proceedings of the 16th International Workshop on Semantic
  Evaluation (SemEval-2022)}, pages 1412--1437, Seattle, United States.
  Association for Computational Linguistics.

\bibitem[{Nan et~al.(2020)Nan, Radev, Zhang, Rau, Sivaprasad, Hsieh, Tang,
  Vyas, Verma, Krishna, Liu, Irwanto, Pan, Rahman, Zaidi, Mutuma, Tarabar,
  Gupta, Yu, Tan, Lin, Xiong, Socher, and Rajani}]{DART}
Linyong Nan, Dragomir Radev, Rui Zhang, Amrit Rau, Abhinand Sivaprasad,
  Chiachun Hsieh, Xiangru Tang, Aadit Vyas, Neha Verma, Pranav Krishna,
  Yangxiaokang Liu, Nadia Irwanto, Jessica Pan, Faiaz Rahman, Ahmad Zaidi,
  Mutethia Mutuma, Yasin Tarabar, Ankit Gupta, Tao Yu, Yi~Chern Tan,
  Xi~Victoria Lin, Caiming Xiong, Richard Socher, and Nazneen~Fatema Rajani.
  2020.
\newblock \href {https://doi.org/10.48550/arxiv.2007.02871} {{DART: Open-Domain
  Structured Data Record to Text Generation}}.
\newblock \emph{arXiv}.

\bibitem[{Novikova et~al.(2017)Novikova, Du{\v{s}}ek, and
  Rieser}]{novikova-etal-2017-e2e}
Jekaterina Novikova, Ond{\v{r}}ej Du{\v{s}}ek, and Verena Rieser. 2017.
\newblock \href {https://doi.org/10.18653/v1/W17-5525} {The {E}2{E} dataset:
  New challenges for end-to-end generation}.
\newblock In \emph{Proceedings of the 18th Annual {SIG}dial Meeting on
  Discourse and Dialogue}, pages 201--206, Saarbr{\"u}cken, Germany.
  Association for Computational Linguistics.

\bibitem[{Pang and Gimpel(2019)}]{NonParaTextTransfer}
Richard~Yuanzhe Pang and Kevin Gimpel. 2019.
\newblock \href {https://doi.org/10.18653/v1/d19-5614} {{Unsupervised
  Evaluation Metrics and Learning Criteria for Non-Parallel Textual Transfer}}.
\newblock \emph{Proceedings of the 3rd Workshop on Neural Generation and
  Translation}, pages 138--147.

\bibitem[{Papineni et~al.(2002)Papineni, Roukos, Ward, and
  Zhu}]{papineni-etal-2002-bleu}
Kishore Papineni, Salim Roukos, Todd Ward, and Wei-Jing Zhu. 2002.
\newblock \href {https://doi.org/10.3115/1073083.1073135} {{B}leu: a method for
  automatic evaluation of machine translation}.
\newblock In \emph{Proceedings of the 40th Annual Meeting of the Association
  for Computational Linguistics}, pages 311--318, Philadelphia, Pennsylvania,
  USA. Association for Computational Linguistics.

\bibitem[{Parikh et~al.(2020)Parikh, Wang, Gehrmann, Faruqui, Dhingra, Yang,
  and Das}]{ToTTo}
Ankur Parikh, Xuezhi Wang, Sebastian Gehrmann, Manaal Faruqui, Bhuwan Dhingra,
  Diyi Yang, and Dipanjan Das. 2020.
\newblock \href {https://doi.org/10.18653/v1/2020.emnlp-main.89} {{ToTTo: A
  Controlled Table-To-Text Generation Dataset}}.
\newblock \emph{Proceedings of the 2020 Conference on Empirical Methods in
  Natural Language Processing (EMNLP)}, pages 1173--1186.

\bibitem[{Pasupat and Liang(2015)}]{pasupat-liang-2015-compositional}
Panupong Pasupat and Percy Liang. 2015.
\newblock \href {https://doi.org/10.3115/v1/P15-1142} {Compositional semantic
  parsing on semi-structured tables}.
\newblock In \emph{Proceedings of the 53rd Annual Meeting of the Association
  for Computational Linguistics and the 7th International Joint Conference on
  Natural Language Processing (Volume 1: Long Papers)}, pages 1470--1480,
  Beijing, China. Association for Computational Linguistics.

\bibitem[{Raffel et~al.(2020)Raffel, Shazeer, Roberts, Lee, Narang, Matena,
  Zhou, Li, and Liu}]{t5}
Colin Raffel, Noam Shazeer, Adam Roberts, Katherine Lee, Sharan Narang, Michael
  Matena, Yanqi Zhou, Wei Li, and Peter~J. Liu. 2020.
\newblock Exploring the limits of transfer learning with a unified text-to-text
  transformer.
\newblock \emph{J. Mach. Learn. Res.}, 21(1).

\bibitem[{Su et~al.(2021)Su, Meng, Baker, and
  Collier}]{su-etal-2021-shot-table}
Yixuan Su, Zaiqiao Meng, Simon Baker, and Nigel Collier. 2021.
\newblock \href {https://doi.org/10.18653/v1/2021.findings-emnlp.77} {Few-shot
  table-to-text generation with prototype memory}.
\newblock In \emph{Findings of the Association for Computational Linguistics:
  EMNLP 2021}, pages 910--917, Punta Cana, Dominican Republic. Association for
  Computational Linguistics.

\bibitem[{Vedula et~al.(2022)Vedula, Collins, Agichtein, and
  Rokhlenko}]{vedula2022matters}
Nikhita Vedula, Marcus Collins, Eugene Agichtein, and Oleg Rokhlenko. 2022.
\newblock What matters for shoppers: Investigating key attributes for online
  product comparison.
\newblock In \emph{European Conference on Information Retrieval}, pages
  231--239. Springer.

\bibitem[{Vedula et~al.(2023)Vedula, Collins, Agichtein, and
  Rokhlenko}]{vedula2023generating}
Nikhita Vedula, Marcus Collins, Eugene Agichtein, and Oleg Rokhlenko. 2023.
\newblock Generating explainable product comparisons for online shopping.
\newblock In \emph{Proceedings of the Sixteenth ACM International Conference on
  Web Search and Data Mining}, pages 949--957.

\bibitem[{Williams and Zipser(1989)}]{williams1989learning}
Ronald~J Williams and David Zipser. 1989.
\newblock A learning algorithm for continually running fully recurrent neural
  networks.
\newblock \emph{Neural computation}, 1(2):270--280.

\bibitem[{Xiang et~al.(2022)Xiang, Liu, Zhou, Xing, and
  Hu}]{xiang-etal-2022-asdot}
Jiannan Xiang, Zhengzhong Liu, Yucheng Zhou, Eric Xing, and Zhiting Hu. 2022.
\newblock \href {https://aclanthology.org/2022.findings-emnlp.136} {{ASDOT}:
  Any-shot data-to-text generation with pretrained language models}.
\newblock In \emph{Findings of the Association for Computational Linguistics:
  EMNLP 2022}, pages 1886--1899, Abu Dhabi, United Arab Emirates. Association
  for Computational Linguistics.

\bibitem[{Yang et~al.(2022)Yang, Gupta, Upadhyay, He, Goel, and
  Paul}]{TableFormer}
Jingfeng Yang, Aditya Gupta, Shyam Upadhyay, Luheng He, Rahul Goel, and Shachi
  Paul. 2022.
\newblock \href {http://arxiv.org/abs/2203.00274} {{TableFormer: Robust
  Transformer Modeling for Table-Text Encoding}}.
\newblock \emph{arXiv}.
\newblock Very interesting approach to use scalar attention biases between
  different types of content, e.g. table columns and the input query.

\bibitem[{Zhang et~al.(2020)Zhang, Kishore, Wu, Weinberger, and
  Artzi}]{bert-score}
Tianyi Zhang, Varsha Kishore, Felix Wu, Kilian~Q. Weinberger, and Yoav Artzi.
  2020.
\newblock \href {https://openreview.net/forum?id=SkeHuCVFDr} {Bertscore:
  Evaluating text generation with bert}.
\newblock In \emph{International Conference on Learning Representations}.

\bibitem[{Zhong et~al.(2017)Zhong, Xiong, and Socher}]{zhong2017seq2sql}
Victor Zhong, Caiming Xiong, and Richard Socher. 2017.
\newblock Seq2sql: Generating structured queries from natural language using
  reinforcement learning.
\newblock \emph{arXiv preprint arXiv:1709.00103}.

\bibitem[{Zhou et~al.(2016)Zhou, Krähenbühl, Aubry, Huang, and
  Efros}]{10.1109/cvpr.2016.20}
Tinghui Zhou, Philipp Krähenbühl, Mathieu Aubry, Qixing Huang, and Alexei~A.
  Efros. 2016.
\newblock \href {https://doi.org/10.1109/cvpr.2016.20} {{Learning Dense
  Correspondence via 3D-Guided Cycle Consistency}}.
\newblock \emph{2016 IEEE Conference on Computer Vision and Pattern Recognition
  (CVPR)}, pages 117--126.

\bibitem[{Zhu et~al.(2017)Zhu, Park, Isola, and Efros}]{UnpairedImageImage}
Jun-Yan Zhu, Taesung Park, Phillip Isola, and Alexei~A. Efros. 2017.
\newblock \href {https://doi.org/10.1109/iccv.2017.244} {{Unpaired
  Image-to-Image Translation Using Cycle-Consistent Adversarial Networks}}.
\newblock \emph{2017 IEEE International Conference on Computer Vision (ICCV)},
  pages 2242--2251.

\end{thebibliography}
\bibliographystyle{acl_natbib}

\clearpage

\appendix
\section*{\centering Appendix}

\renewcommand{\arraystretch}{1.1}

\section{Annotation Guidelines}
\label{sec:guidelines}
In this section, we include descriptions of the human annotation task performed in this work.

For this annotation task, the annotators will be provided a set of input triplets in the subject-predicate-object structure, and the annotators will be asked to provide their judgement of four model-generated text snippets associated with the input triplets. 
Our target is to annotate the 1) Count of Factual Errors, 2) Count of Hallucination Errors, 3) Count of Information Misses, and 4) Fluency Preference for the generations. We use two different Appen interface-pages: one for the annotation of the three types of error counts, and one for the annotation of Fluency Preference. 

\subsection{Annotation of Error Counts}
\subsubsection{Count of Factual Errors (FE)}
Count of Factual Errors (FE) measures the factual correctness of the generated text with respect to the entities (subject and object) and predicates of the input triplets. 

\noindent\textbf{Annotation Instruction:} Factual errors are information in the generations which contradict the information in the subject-predictate-object context. 
For each attempted predicate given in the input triplets, the annotator is supposed to increase the count if [the subject and/or object of the predicate’s associated expression \emph{does \textbf{not} match the facts} suggested by the input triplets]. 

\noindent\textbf{Examples:} (See Table \ref{tab:fe_examples})

\begin{table*}
\footnotesize
\centering
\begin{tabular}{|c|p{13.5cm}|} \hline
\begin{tabular}[c]{@{}c@{}}Input Triple\\Set 1\end{tabular}          & \begin{tabular}[c]{@{}p{13.5cm}@{}}1. [S] Mexico [P] currency [O] Mexican peso\\2. [S] Mexico [P] demonym [O] Mexicans\\3. [S] Bionico [P] course [O] Dessert\\4. [S] Bionico [P] ingredient [O] Raisin\\5. [S] Bionico [P] country [O] Mexico\end{tabular}\\ \hline
\begin{tabular}[c]{@{}c@{}}Generations\\and Reasonings\end{tabular}  & \begin{tabular}{@{}p{13.5cm}@{}}{\labelitemi}\hspace{\dimexpr\labelsep+0.5\tabcolsep}1 FE: Bionico is a dessert made with Raisin~\textcolor[rgb]{1,0.212,0.2}{and Mexican peso}. It is a dish from Mexico.\\\hspace{0.5\leftmargin}{\labelitemii}\hspace{\dimexpr\labelsep+0.5\tabcolsep}According to the input data, Mexican peso is the currency of Mexico not the ingredient of Bionico, so it is a FE.\\{\labelitemi}\hspace{\dimexpr\labelsep+0.5\tabcolsep}\textcolor[rgb]{0.2,0.2,0.2}{2 FEs:~In Mexico, the currency is the Mexican peso.}\textcolor[rgb]{1,0.212,0.2}{~It is a dessert with a Raisin ingredient}\textcolor[rgb]{0.2,0.2,0.2}{.}\\\hspace{0.5\leftmargin}{\labelitemii}\hspace{\dimexpr\labelsep+0.5\tabcolsep}\textcolor[rgb]{0.2,0.2,0.2}{"It" is a pronoun that grammatically refers to Mexican peso, so the subjects of attempted expressions for triplet }3 and 4 are wrong, which results in two FEs.\\{\labelitemi}\hspace{\dimexpr\labelsep+0.5\tabcolsep}1 FE: Bionico is the~\textcolor[rgb]{0.2,0.2,0.2}{demonym of~Raisin}\\\hspace{0.5\leftmargin}{\labelitemii}\hspace{\dimexpr\labelsep+0.5\tabcolsep}This is considered as an attempt to express triplet 2 but is factually incorrect.~\end{tabular}\\ \hline
\begin{tabular}[c]{@{}c@{}}Input Triple\\Set 2\end{tabular}          & \begin{tabular}[c]{@{}p{13.5cm}@{}}1.~[S] Alan B. Miller Hall [P] address [O] 101 Ukrop Way\\2.~[S] Alan B. Miller Hall [P] height [O] 36.5 meters\end{tabular}\\ \hline
\begin{tabular}[c]{@{}c@{}}Generations\\and Reasonings\end{tabular}  & \begin{tabular}{@{}p{13.5cm}@{}}{\labelitemi}\hspace{\dimexpr\labelsep+0.5\tabcolsep}\textcolor[rgb]{0.2,0.2,0.2}{2 FEs: Alan B. Miller Hall located at~}\textcolor[rgb]{1,0.212,0.2}{440 Terry Avenue}\textcolor[rgb]{0.2,0.2,0.2}{~has a height of~}\textcolor[rgb]{1,0.212,0.2}{365}\textcolor[rgb]{0.2,0.2,0.2}{~meters.}\\\hspace{0.5\leftmargin}{\labelitemii}\hspace{\dimexpr\labelsep+0.5\tabcolsep}Although~\textit{440 Terry Avenue}~and~\textit{365}~may seem like hallucinations, they counter the fact that the address of Alan B. Miller Hall is 101 Ukrop Way and the fact that the Hall’s height is 36.5 meters. We consider them as FEs instead of HEs because the input data explicitly contradicts these generated strings (which is how FEs are defined).~\end{tabular}\\ \hline
\end{tabular}
\caption{Disambiguation examples of Factual Errors (FE).}
\label{tab:fe_examples}
\end{table*}

\subsubsection{Count of Hallucination Errors (HE)}
Count of Hallucination Errors (HE) measures the relevance of the generated text with respect to the input triplets. 

\noindent\textbf{Annotation Instruction:} Hallucination errors occur when words or phrases in the generation cannot be inferred from the subject-predicate-object triplets, for instance because the value doesn’t make logical sense, or because the predicate of the expression isn’t present in any triple. Distinguished from FEs, HEs invent information not in the triplets or reference, but do not directly contradict the triplets.
The annotator is supposed to increase the count if [a piece of information contained in the generated text \emph{is \textbf{not} presented in} or \emph{can \textbf{not} be reasonably inferred by} the input triplets]. For better consistency and less ambiguity, reasonable inference is defined as a piece of information contained in the generated text isn’t presented in the input triplets but is presented in the reference text. 

\noindent\textbf{Examples:} (See Table \ref{tab:he_examples})

\begin{table*}
\footnotesize
\centering
\begin{tabular}{|c|p{13.5cm}|} \hline
\multicolumn{2}{|c|}{\textbf{Count of Hallucination Errors (HE)}}\\ \hline
\begin{tabular}[c]{@{}c@{}}Input Triple\\Set 1\end{tabular}          & \begin{tabular}[c]{@{}p{13.5cm}@{}}1. [S] ALCO RS-3 [P] build date [O] May 1950 - August 1956\\2. [S] ALCO RS-3 [P] power type [O] Diesel-electric transmission\\3.~[S] ALCO RS-3 [P] builder [O] Montreal Locomotive Works\\4.~[S] ALCO RS-3 [P] length [O] 17068.8\end{tabular}\\ \hline
Reference Text                                                       & \begin{tabular}[c]{@{}p{13.5cm}@{}}-~The ALCO RS-3 was produced between May 1950 and August 1956 and was built by Montreal Locomotive Works. This locomotive has a diesel-electric transmission and is 17068.8 millimetres in length.\\-~The ALCO RS-3 was produced between May 1950 and August 1956 and was built by Montreal Locomotive Works. It has a diesel-electric transmission and is 17068.8 millimetres long.\\-~The ALCO RS-3, built by the Montreal Locomotive Works between May 1950 and August 1956, has a diesel-electric transmission and measures 17068.8 millimetres in length.\end{tabular}\\ \hline
\begin{tabular}[c]{@{}c@{}}Generations~\\and Reasonings\end{tabular} & \begin{tabular}{@{}p{13.5cm}@{}}{\labelitemi}\hspace{\dimexpr\labelsep+0.5\tabcolsep}1 HE: The Montreal Locomotive Works built the ALCO RS-3 from May 1950 - August 1956. It has a diesel-electric transmission and a length of 17068.8~\textcolor{red}{meters}.\\\hspace{0.5\leftmargin}{\labelitemii}\hspace{\dimexpr\labelsep+0.5\tabcolsep}The unit expression of \textit{meters} is considered as a HE since such information doesn't appear in the input data or the reference text (hence not considered as a reasonable inference).\\{\labelitemi}\hspace{\dimexpr\labelsep+0.5\tabcolsep}0 HE: The ALCO RS-3 was built by the Montreal Locomotive Works between May 1950 and August 1956. It has a diesel-electric transmission and is 17068.8 \textcolor[rgb]{0.98,0.733,0.016}{millimetres} long.\\\hspace{0.5\leftmargin}{\labelitemii}\hspace{\dimexpr\labelsep+0.5\tabcolsep}The unit expression of~\textit{milimeters}~doesn’t appear in the input data but appears in the reference text (hence it is considered as a reasonable~inference), so it is not a HE.\end{tabular}\\ \hline
\begin{tabular}[c]{@{}c@{}}Input Triple\\Set 2\end{tabular}          & \begin{tabular}[c]{@{}p{13.5cm}@{}}1.~[S] Liselotte Grschebina [P] death place [O] Israel\\2.~[S] Liselotte Grschebina [P] death place [O] Petah Tikva\\3.~[S] Israel [P] population density [O] 387.63\\4.~[S] Israel [P] long name [O] State of Israel\\5.~[S] Liselotte Grschebina [P] nationality [O] Israel\end{tabular}\\ \hline
Reference Text                                                       & \begin{tabular}[c]{@{}p{13.5cm}@{}}-~Liselotte Grschebina is an Israeli national who died in Petah Tikva, Israel which is formally known as the State of Israel and has a population density of 387.63 people per square kilometre of land area.\\-~Liselotte Grschebina was an Israeli who died in Petah Tikva, Israel which has a population density of 387.63 people per square kilometre of land area and is named "State of Israel."\\-~Liselotte Grschebina has Israeli nationality and died in Petah Tikva, Israel. Israel has the population density of 387.63 and its full name is the State of Israel.\end{tabular}\\ \hline
\begin{tabular}[c]{@{}c@{}}Generations~\\and Reasonings\end{tabular} & \begin{tabular}{@{}p{13.5cm}@{}}{\labelitemi}\hspace{\dimexpr\labelsep+0.5\tabcolsep}1 HE:~\textcolor{red}{Liselotte Grschebina was born in Israel}~and died in Petah Tikva. Israel has a population density of 387.63 people\textit{.}\\\hspace{0.5\leftmargin}{\labelitemii}\hspace{\dimexpr\labelsep+0.5\tabcolsep}The birth place information doesn’t appear in the input data and cannot be reasonably inferred either, so it is considered as a HE.~\end{tabular}\\ \hline
\end{tabular}
\caption{Disambiguation examples of Hallucination Errors (HE).}
\label{tab:he_examples}
\end{table*}

\subsubsection{Count of Information Misses (IM)}
Count of Information Misses (IM) measures the information coverage of the generated text with respect to the predicates given in the input triplets.

\noindent\textbf{Annotation Instruction:} For each predicate given in the input triplets, the annotator is supposed to increase the count by 1 if [the generated text \emph{did \textbf{not} attempt} to express the predicate].

\noindent\textbf{Examples:} (See Table \ref{tab:im_examples})

\begin{table*}
\footnotesize
\centering
\begin{tabular}{|c|p{13.5cm}|} \hline
\begin{tabular}[c]{@{}c@{}}Input Triple\\Set 1\end{tabular}          & \begin{tabular}[c]{@{}p{13.5cm}@{}}1.~\textcolor[rgb]{0.749,0.039,0.761}{[S] Liselotte Grschebina [P] birth place [O] Karlsruhe}\\2.~\textcolor[rgb]{0.353,0.039,0.761}{[S] Liselotte Grschebina [P] nationality [O] Israel}3.~\textcolor[rgb]{0.039,0.329,0.761}{[S] Liselotte Grschebina [P] training [O] School of Applied Arts in Stuttgart}\\4.~\textcolor[rgb]{0.173,0.671,0.129}{[S] Karlsruhe [P] country [O] Germany}\\5.~\textcolor[rgb]{0.8,0.722,0}{[S] Israel [P] language [O] Modern Hebrew}\end{tabular}\\ \hline
\begin{tabular}[c]{@{}c@{}}Generations~\\and Reasonings\end{tabular} & \begin{tabular}{@{}p{13.5cm}@{}}{\labelitemi}\hspace{\dimexpr\labelsep+0.5\tabcolsep}1 IM:~\textcolor[rgb]{0.749,0.039,0.761}{Liselotte Grschebina was born in~}\textcolor[rgb]{0.749,0.039,0.761}{Karlsruhe},~\textcolor[rgb]{0.173,0.671,0.129}{Germany}.~\textcolor[rgb]{0.039,0.329,0.761}{She studied at the School of Applied Arts in Stuttgart}.~\textcolor[rgb]{0.8,0.722,0}{Modern Hebrew is spoken in Israel.}\\\hspace{0.5\leftmargin}{\labelitemii}\hspace{\dimexpr\labelsep+0.5\tabcolsep}Triplet 2 hasn’t been expressed.\\\hspace{0.5\leftmargin}{\labelitemii}\hspace{\dimexpr\labelsep+0.5\tabcolsep}\textbf{The expression of a predicate can be implicit.}~For instance,~\textit{Karlsruhe, Germany}~is an implicit expression with respect to~\textit{triplet 4}.\\{\labelitemi}\hspace{\dimexpr\labelsep+0.5\tabcolsep}2 IMs:~\textcolor[rgb]{0.749,0.039,0.761}{Liselotte Grschebina was born in Karlsruhe},~Israel~and~\textcolor[rgb]{0.039,0.329,0.761}{trained at the School of Applied Arts in Stuttgart.}\\\hspace{0.5\leftmargin}{\labelitemii}\hspace{\dimexpr\labelsep+0.5\tabcolsep}Triple 2 and 5 haven’t been expressed.\\\hspace{0.5\leftmargin}{\labelitemii}\hspace{\dimexpr\labelsep+0.5\tabcolsep}\textit{Karlsruhe},~\textit{Israel~}can be considered as an expression attempt of triplet 4 although it contains factual errors. IM only counts information coverage with respect to the predicates and neglects entities (subject/object).\\{\labelitemi}\hspace{\dimexpr\labelsep+0.5\tabcolsep}0 IM:~\textcolor[rgb]{0.749,0.039,0.761}{Liselotte Grschebina was born in Karlsruhe,}~\textcolor[rgb]{0.173,0.671,0.129}{Germany}~\textcolor[rgb]{0.039,0.329,0.761}{and studied at the School of Applied Arts in Stuttgart.}~\textcolor[rgb]{0.353,0.039,0.761}{She is Israeli}~\textcolor[rgb]{0.8,0.722,0}{and speaks Modern Hebrew.}\\\hspace{0.5\leftmargin}{\labelitemii}\hspace{\dimexpr\labelsep+0.5\tabcolsep}\textit{(She/Liselotte) speaks Modern Hebrew}~can be considered as an~expression attempt of triplet 5.~\textit{Somebody(Israeli) speaks Modern Hebrew}~is a reasonable alternative expression attempt of~\textit{the language in Israel is Modern Hebrew}.~\end{tabular}  \\ \hline
\begin{tabular}[c]{@{}c@{}}Input Triple\\Set 2\end{tabular}          & \begin{tabular}[c]{@{}p{13.5cm}@{}}1.~\textcolor[rgb]{0.749,0.039,0.761}{[S] Liselotte Grschebina [P] death place [O] Israel}\\2.~\textcolor[rgb]{0.353,0.039,0.761}{[S] Liselotte Grschebina [P] death place [O] Petah Tikva}\end{tabular}\\ \hline
\begin{tabular}[c]{@{}c@{}}Generations\\and Reasonings\end{tabular}  & \begin{tabular}{@{}p{13.5cm}@{}}{\labelitemi}\hspace{\dimexpr\labelsep+0.5\tabcolsep}1 IM: Liselotte Grschebina died in Petah Tikva.\\\hspace{0.5\leftmargin}{\labelitemii}\hspace{\dimexpr\labelsep+0.5\tabcolsep}This is a special case which we count as having a IM. In rare cases, the predicates in the input data may look the same due to~omissions. Here, the predicate of triplet 1 is actually~\textit{death place (country)}~and of triplet 2 is actually~\textit{death place (city)}. Hence, this generation only expresses one triplet’s predicate.\end{tabular}\\ \hline
\end{tabular}
\caption{Disambiguation examples of Information Misses (IM).}
\label{tab:im_examples}
\end{table*}

\subsubsection{Annotation Interface for Errors}

\begin{figure*}[t]
\centering
\includegraphics[width=0.8\textwidth]{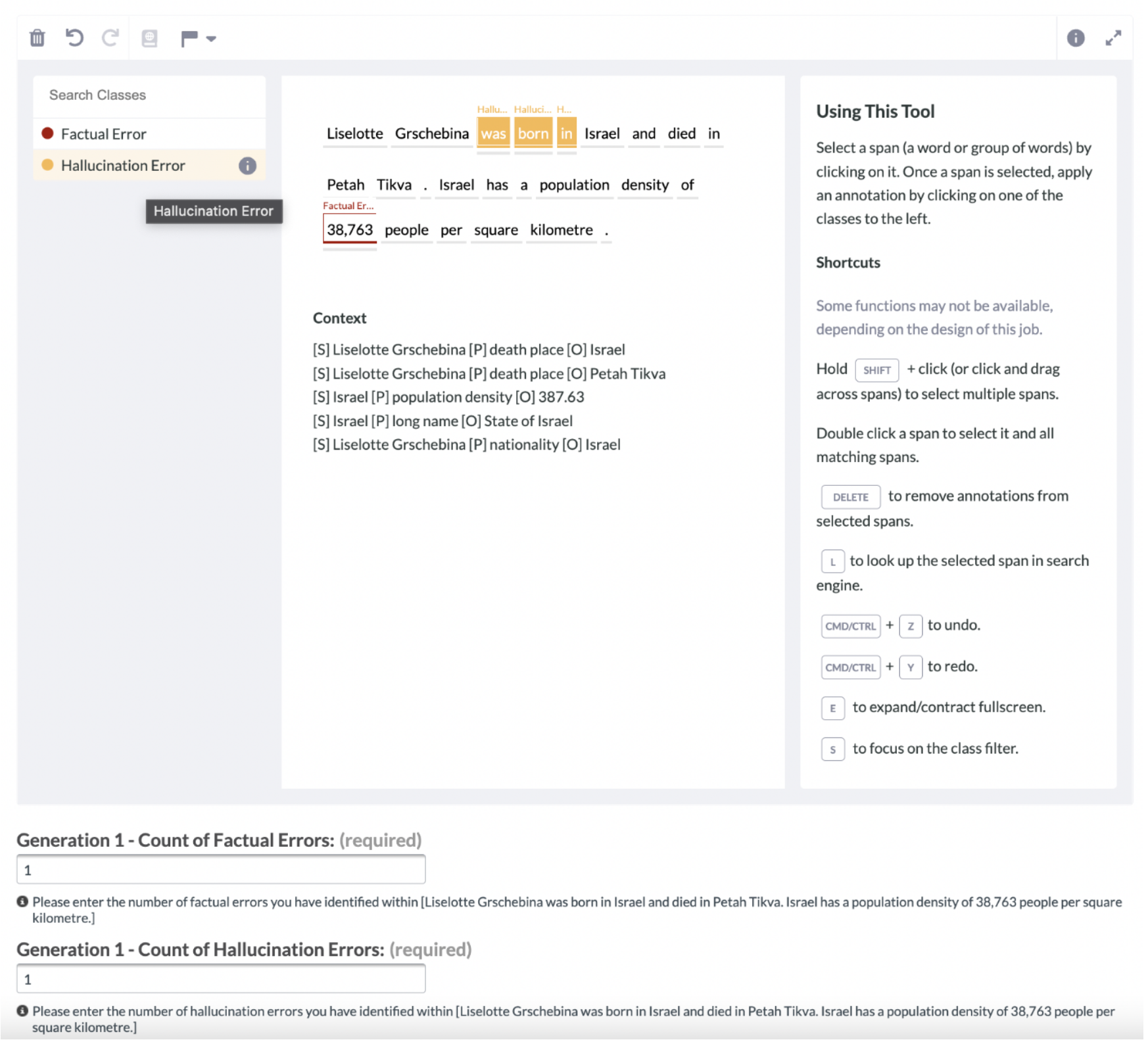}
\caption{Annotation Interface for FE and HE.}
\label{interface_1}
\end{figure*}

\begin{figure*}[t]
\centering
\includegraphics[width=0.8\textwidth]{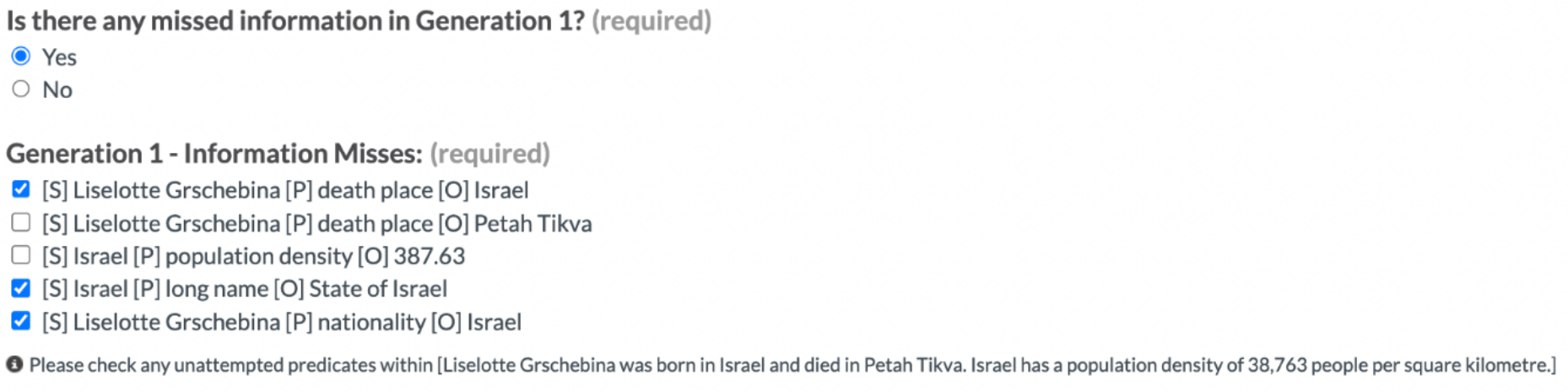}
\caption{Annotation Interface for IM.}
\label{interface_2}
\end{figure*}

The annotation task is presented batch-by-batch. Each batch contains one shared input triplet and three model-generated text snippets (in random order) with respect to the input triplets. The annotators will see the input triplets data and the reference ground-truth data at first. Please keep in mind that the ground-truth data is just a reference for the convenience of better understanding the input triplets and the boundary of "reasonable inference" and they may not be perfect. To begin with, we ask the annotators to provide token level annotations of FE and HE. The “Context” is the input triplets shown before. The annotators can click the [ grey-rounded i ] button at the upper-right conner to see information regarding the use of the annotation tool. The annotators can also click the [grey-rounded i] button next to the tag to see a recap of its definition. Annotations of overlapped tokens are permitted. After finishing up the token-level FE and HE annotation, please provide the count of FE and the count of HE respectively. Next, the annotators need to identify if there’s any missed information in the generation. If "Yes", the annotators will be asked to check the IMs.
See Figure \ref{interface_1} and Figure \ref{interface_2} for screenshots of the annotation interface for FE, HE, and IM.

\subsubsection{Fluency Preference (FP)}
Fluency Preference (FP) measures the quality of the generated text in terms of the grammar, structure, and the coherence of the text.

\noindent\textbf{Annotation Instruction:} The annotator is supposed to perform pairwise fluency comparison of the generated texts within a batch to compile the final ranking that reflects the annotator’s subjective preference. The fluency comparison and ranking shall only consider the \emph{grammar}, \emph{structure}, and the \emph{coherence} of the text \textbf{without} the consideration of IM, FE, and HE.

\noindent\textbf{Examples:} Since FP is a relatively more subjective measure that asks for overall preference, we only provide some contrasting examples for the three aspects of fluency. 
\begin{itemize}
    \item \underline{Grammar}: Generation A is better than B because B is grammatically incorrect/influent.
    \begin{itemize}
        \item Generation A: 108, written by karen maser, has 2.12 million U.S. viewers.
        \item Generation B: 108 U.S. viewers million is 2.12, written by karen maser.
    \end{itemize}
    \item \underline{Structure}: Generation A is better than B because the pieces of information in A are more naturally connected and expressed.
    \begin{itemize}
        \item Generation A: Andrew Rayel is a member of the Bobina band that plays trance music.
        \item Generation B: Andrew Rayel is an associated band/associated musical artist with Bobina. His genre is Trance music.
    \end{itemize}
    \item \underline{Coherence}: Generation A is better than B because \emph{She speaks modern Hebrew} is more logically and consistently connected with the previous sentences compared to \emph{Modern Hebrew is spoken in Israel}. 
    \begin{itemize}
        \item Generation A: Liselotte Grschebina was born in Karlsruhe, Germany and trained in the School of Applied Arts in Stuttgart. She speaks modern Hebrew.
        \item Generation B: Liselotte Grschebina was born in Karlsruhe, Germany. She studied at the School of Applied Arts in Stuttgart. Modern Hebrew is spoken in Israel.
    \end{itemize}
\end{itemize}

\subsubsection{Annotation Interface for FP}
The annotators may see two to three generations, and the annotators are asked to perform pairwise comparison and rank the generations by their grammar, structure, and coherence without considering information coverage and factual errors. The annotators should start with 1 for the highest-ranked/most-fluent text of the generations within the batch. Ranking tie is permitted, but note this is a ranking task, so the annotators will need to check the numbers in a normal ranking manner. If the annotators see two generations [A, B], and A is better than B, then the annotators should select 1 for A and 2 for B instead of 3 for B. If the annotators see three generations [A, B, C], and A is identical to B, B is better than C, then the annotators should select 1 for A, 1 for B, 3 for C instead of 2 for C. See Fi-
\begin{figure*}[t]
\centering
\includegraphics[width=0.8\textwidth]{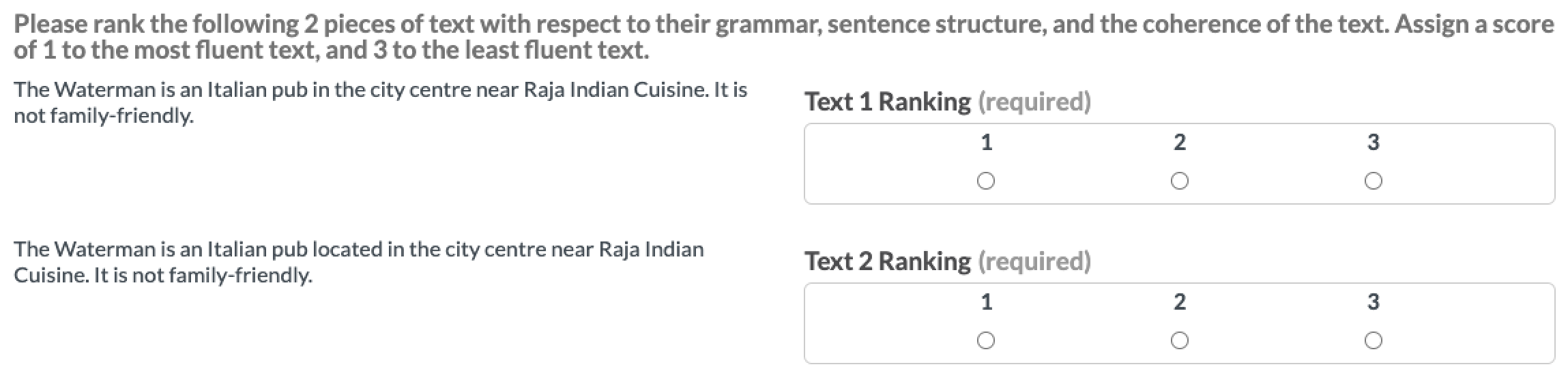}
\caption{Annotation Interface for FP.}
\label{interface_3}
\end{figure*}

\clearpage

\noindent -gure \ref{interface_3} for a screenshot of the annotation interface for FP.

\section{Additional Result Analysis for WebNLG and DART}
\label{sec:appendix}

In this section, we show some additional samples of text generated by multiple models trained under different cycle settings, and the errors made by these models for the WebNLG dataset in Table \ref{tab:appendix_table1} and DART dataset in Table \ref{tab:appendix_table2}.

\begin{table*}
\footnotesize
\centering
\begin{tabular}{|c|l|c|} 
\hline
\multicolumn{3}{|c|}{\textbf{WebNLG - Sample 1}}                                                                                                                                                                                                                                                                                                                                                                                                                                                                                                                                                                                                                                                                                                \\ 
\hline
\textbf{Input Triple Set}    & \multicolumn{2}{l|}{\begin{tabular}[c]{@{}l@{}}1. \textcolor[rgb]{0.259,0.522,0.957}{[S] Liselotte Grschebina [P] death place [O] Israel}\\2. \textcolor[rgb]{0.918,0.263,0.208}{[S] Liselotte Grschebina [P] death place [O] Petah Tikva}\\3. \textcolor[rgb]{0.984,0.737,0.016}{[S] Israel [P] population density [O] 387.63}\\4. \textcolor[rgb]{0.204,0.659,0.325}{[S] Israel [P] long name [O] State of Israel}\\5. \textcolor[rgb]{1,0.427,0.004}{[S] Liselotte Grschebina [P] nationality [O] Israel}\end{tabular}}                                                                                                                                                                                       \\ 
\hline
\textbf{Reference Text}      & \multicolumn{2}{l|}{\begin{tabular}[c]{@{}l@{}}- Liselotte Grschebina is an Israeli national who died in Petah Tikva, Israel which is formally known as\\the State of Israel and has a population density of 387.63 people per square kilometre of land area.\\- Liselotte Grschebina was an Israeli who died in Petah Tikva, Israel which has a population density of\\387.63 people per square kilometre of land area and is named "State of Israel".\\- Liselotte Grschebina has Israeli nationality and died in Petah Tikva, Israel. Israel has the population\\density of 387.63 and its full name is the State of Israel.\end{tabular}}                                                                    \\ 
\hline
\textbf{Method}              & \multicolumn{1}{c|}{\textbf{Generation}}                                                                                                                                                                                                                                                                                                                                                                                                                                                                                      & \textbf{Remark}                                                                                                                                                                  \\ 
\hline
\begin{tabular}[c]{@{}c@{}}Fully-supervised\\fine-tuning\end{tabular} & \begin{tabular}[c]{@{}l@{}}\textcolor[rgb]{0.259,0.522,0.957}{Liselotte Grschebina was an Israeli national} \textcolor[rgb]{0.918,0.263,0.208}{who died in Petah Tikva}\textcolor[rgb]{0.259,0.522,0.957}{, Israel.} \textcolor[rgb]{0.204,0.659,0.325}{The long}\\\textcolor[rgb]{0.204,0.659,0.325}{name of Israel is the State of Israel} \textcolor[rgb]{0.984,0.737,0.016}{and the population density is 387.63 inhabitants per}\\\textcolor[rgb]{0.984,0.737,0.016}{square kilometre.}\end{tabular}                     &                                                                                                                                                                                  \\ 
\hline
\begin{tabular}[c]{@{}c@{}}Low-resource\\fine-tuning\end{tabular}     & \begin{tabular}[c]{@{}l@{}}Liselotte Grschebina was born in Israel \textcolor[rgb]{0.918,0.263,0.208}{and died in Petah Tikva.} \textcolor[rgb]{0.259,0.522,0.957}{Israel has a population}\\\textcolor[rgb]{0.259,0.522,0.957}{density of }\textcolor[rgb]{0.259,0.522,0.957}{387,63 people}\textcolor[rgb]{0.259,0.522,0.957}{.}\end{tabular}                                                                                                                                                                               & \begin{tabular}[c]{@{}c@{}}3 IMs,\\1 FE, 1HE\end{tabular}                                                                                                                        \\ 
\hline
\begin{tabular}[c]{@{}c@{}}Low-resource\\cycle-training\end{tabular}  & \begin{tabular}[c]{@{}l@{}}\textcolor[rgb]{0.918,0.263,0.208}{Liselotte Grschebina died in Petah Tikva}\textcolor[rgb]{0.259,0.522,0.957}{, Israel.} \textcolor[rgb]{0.984,0.737,0.016}{The population density of Israel is 387.63}\\\textcolor[rgb]{0.984,0.737,0.016}{inhabitants per square kilometre }\textcolor[rgb]{0.204,0.659,0.325}{and the long name is the State of Israel.}\end{tabular}                                                                                                                          & 1 IM                                                                                                                                                                             \\ 
\hline
\multicolumn{3}{|c|}{\textbf{WebNLG - Sample 2}}                                                                                                                                                                                                                                                                                                                                                                                                                                                                                                                                                                                                                                                                                                \\ 
\hline
\textbf{Input Triple Set}    & \multicolumn{2}{l|}{\begin{tabular}[c]{@{}l@{}}1. \textcolor[rgb]{0.259,0.522,0.957}{[S] Alan B. Miller Hall [P] architect [O] Robert A. M. Stern}\\2. \textcolor[rgb]{0.918,0.263,0.208}{[S] Alan B. Miller Hall [P] address [O] 101 Ukrop Way}\\3. \textcolor[rgb]{0.984,0.737,0.016}{[S] Alan B. Miller Hall [P] current tenants [O] Mason School of Business}\\4. \textcolor[rgb]{0.204,0.659,0.325}{[S] Alan B. Miller Hall [P] completion date [O] 2009-06-01}\\5. \textcolor[rgb]{1,0.427,0.004}{[S] Alan B. Miller Hall [P] location [O] Virginia}\end{tabular}}                                                                                                                                         \\ 
\hline
\textbf{Reference Text}      & \multicolumn{2}{l|}{\begin{tabular}[c]{@{}l@{}}- The Mason School of Business is located at the Alan B. Miller Hall in Virginia at 101 Ukrop Way. The\\architect of the building was Robert A M Stern and the building completion date was 01/06/2007.\\- The address of Alan B. Miller Hall is 101 Ukrop Way, Virginia and the hall is currently tenanted by The\\Mason School of Business. The hall was designed by Robert A.M. Stern and was completed on June\\1st, 2009.\\- Alan B. Miller Hall, which was designed by Robert A.M. Stern is in the State of Virginia, at 101 Ukrop\\Way. The building was finished on 01/06/2009 and is currently tenanted by The Mason School of\\Business.\end{tabular}}  \\ 
\hline
\textbf{Method}              & \multicolumn{1}{c|}{\textbf{Generation}}                                                                                                                                                                                                                                                                                                                                                                                                                                                                                      & \textbf{Remark}                                                                                                                                                                  \\ 
\hline
\begin{tabular}[c]{@{}c@{}}Fully-supervised\\fine-tuning\end{tabular} & \begin{tabular}[c]{@{}l@{}}\textcolor[rgb]{0.984,0.737,0.016}{The Mason School of Business are the current tenants of Alan B Miller Hall} \textcolor[rgb]{0.259,0.522,0.957}{which was}\\\textcolor[rgb]{0.259,0.522,0.957}{designed by Robert A M Stern }\textcolor[rgb]{0.204,0.659,0.325}{and completed on 1 June 2009} \textcolor[rgb]{0.918,0.263,0.208}{at 101 Ukrop Way}\textcolor[rgb]{1,0.427,0.004}{, Virginia.}\end{tabular}                                                                                       &                                                                                                                                                                                  \\ 
\hline
\begin{tabular}[c]{@{}c@{}}Low-resource\\fine-tuning\end{tabular}     & \begin{tabular}[c]{@{}l@{}}\textcolor[rgb]{0.204,0.659,0.325}{Alan B. Miller Hall was completed on June 1, 2009 }\textcolor[rgb]{1,0.427,0.004}{and is located in Virginia.} \textcolor[rgb]{0.259,0.522,0.957}{The architect}\\\textcolor[rgb]{0.259,0.522,0.957}{is Robert A. M. Stern.}\end{tabular}                                                                                                                                                                                                                       & 2 IMs                                                                                                                                                                            \\ 
\hline
\begin{tabular}[c]{@{}c@{}}Low-resource\\cycle-training\end{tabular}  & \begin{tabular}[c]{@{}l@{}}\textcolor[rgb]{0.259,0.522,0.957}{Alan B Miller Hall }\textcolor[rgb]{1,0.427,0.004}{in Virginia}\textcolor[rgb]{0.259,0.522,0.957}{ was designed by Robert A M Stern} \textcolor[rgb]{0.918,0.263,0.208}{and is located at 101}\\\textcolor[rgb]{0.918,0.263,0.208}{Ukrop Way. }\textcolor[rgb]{0.984,0.737,0.016}{The current tenants are the Mason School of Business.} \textcolor[rgb]{0.204,0.659,0.325}{It was completed on 1}\\\textcolor[rgb]{0.204,0.659,0.325}{June 2009.}\end{tabular} &                                                                                                                                                                                  \\ 
\hline
\multicolumn{3}{|c|}{\textbf{WebNLG - Sample 3}}                                                                                                                                                                                                                                                                                                                                                                                                                                                                                                                                                                                                                                                                                                \\ 
\hline
\textbf{Input Triple Set}    & \multicolumn{2}{l|}{\begin{tabular}[c]{@{}l@{}}1. \textcolor[rgb]{0.259,0.522,0.957}{[S] Bootleg Series Volume 1: The Quine Tapes [P] producer [O] The Velvet Underground}\\2. \textcolor[rgb]{0.918,0.263,0.208}{[S] Bootleg Series Volume 1: The Quine Tapes [P] genre [O] Rock music}\\3. \textcolor[rgb]{0.984,0.737,0.016}{[S] The Velvet Underground [P] genre [O] Proto-punk}\end{tabular}}                                                                                                                                                                                                                                                                                                               \\ 
\hline
\textbf{Reference Text}      & \multicolumn{2}{l|}{\begin{tabular}[c]{@{}l@{}}- The rock album "Bootleg Series Volume 1: The Quine Tapes" was produced by proto-punk band The\\Velvet Underground.\\- The Bootleg Series Volume I: The Quine tapes is a rock album, by the proto-punk band, The Velvet\\Underground.\\- Bootleg Series Volume 1: The Quine Tapes is a rock album that also fits into the proto-punk genre. It\\was produced by the Velvet Underground.\end{tabular}}                                                                                                                                                                                                                                                            \\ 
\hline
\textbf{Method}              & \multicolumn{1}{c|}{\textbf{Generation}}                                                                                                                                                                                                                                                                                                                                                                                                                                                                                      & \textbf{Remark}                                                                                                                                                                  \\ 
\hline
\begin{tabular}[c]{@{}c@{}}Fully-supervised\\fine-tuning\end{tabular} & \begin{tabular}[c]{@{}l@{}}\textcolor[rgb]{0.259,0.522,0.957}{The Velvet Underground, }\textcolor[rgb]{0.984,0.737,0.016}{whose genre is proto-punk, }\textcolor[rgb]{0.259,0.522,0.957}{produce the bootleg series volume}\\\textcolor[rgb]{0.259,0.522,0.957}{1: The Quine Tapes.} \textcolor[rgb]{0.918,0.263,0.208}{The genre of the bootleg series is rock music.}\end{tabular}                                                                                                                                          &                                                                                                                                                                                  \\ 
\hline
\begin{tabular}[c]{@{}c@{}}Low-resource\\fine-tuning\end{tabular}     & \begin{tabular}[c]{@{}l@{}}\textcolor[rgb]{0.259,0.522,0.957}{The Velvet Underground is the producer of Bootleg Series Volume 1: The Quine Tapes.}\\\textcolor[rgb]{0.259,0.522,0.957}{}\textcolor[rgb]{0.918,0.263,0.208}{The genre of the bootleg series is rock music.}\end{tabular}
                                                                                                                                                                                                                                                                                                                    & 1 IM                                                                                                                                                                             \\ 
\hline
\begin{tabular}[c]{@{}c@{}}Low-resource\\cycle-training\end{tabular}  & \begin{tabular}[c]{@{}l@{}}\textcolor[rgb]{0.259,0.522,0.957}{The Velvet Underground is the producer of Bootleg Series Volume 1: The Quine Tapes.}\\\textcolor[rgb]{0.259,0.522,0.957}{}\textcolor[rgb]{0.918,0.263,0.208}{The band performs rock music} \textcolor[rgb]{0.984,0.737,0.016}{and their genre is Proto-punk.}\end{tabular}                                                                                                                                                                                      &                                                                                                                                                                                  \\ 
\hline
\end{tabular}
\caption{Error analysis of additional data samples generated by various cycle training settings, for the WebNLG dataset.}
\label{tab:appendix_table1}
\end{table*}

\clearpage

\begin{table*}
\footnotesize
\centering
\begin{tabular}{|c|l|c|} 
\hline
\multicolumn{3}{|c|}{\textbf{DART - Sample 1}}                                                                                                                                                                                                                                                                                                                                                                                                                                                                                                                                                  \\ 
\hline
\textbf{Input Triple Set}    & \multicolumn{2}{l|}{\begin{tabular}[c]{@{}l@{}}1. \textcolor[rgb]{0.259,0.522,0.957}{[S] The Vaults [P] eat type [O] restaurant}\\2. \textcolor[rgb]{0.918,0.263,0.208}{[S] The Vaults [P] food [O] French}\\3. \textcolor[rgb]{0.984,0.737,0.016}{[S] The Vaults [P] pricerange [O] moderate}\\4. \textcolor[rgb]{0.204,0.659,0.325}{[S] The Vaults [P] area [O] riverside}\\5. \textcolor[rgb]{1,0.427,0.004}{[S] The Vaults [P] family friendly [O] yes}\\6. \textcolor[rgb]{0.275,0.741,0.776}{[S] The Vaults [P] near [O] Raja Indian Cuisine}\end{tabular}}  \\ 
\hline
\textbf{Reference Text}      & \multicolumn{2}{l|}{\begin{tabular}[c]{@{}l@{}}- The French restaurant, The Vaults, is a moderately priced, kid friendly restaurant in riverside near Raja\\ Indian Cuisine.\end{tabular}}                                                                                                                                                                                                                                                                                                                                                                        \\ 
\hline
\textbf{Method}              & \multicolumn{1}{c|}{\textbf{Generation}}                                                                                                                                                                                                                                                                                                                                                               & \textbf{Remark}                                                                                                                                         \\ 
\hline
\begin{tabular}[c]{@{}c@{}}Fully-supervised\\fine-tuning\end{tabular} & \begin{tabular}[c]{@{}l@{}}\textcolor[rgb]{0.984,0.737,0.016}{The Vaults is a moderately priced} \textcolor[rgb]{0.918,0.263,0.208}{French restaurant}\textcolor[rgb]{0.204,0.659,0.325}{ in the riverside area }\textcolor[rgb]{0.275,0.741,0.776}{near Raja Indian}\\\textcolor[rgb]{0.275,0.741,0.776}{Cuisine.} \textcolor[rgb]{1,0.427,0.004}{It is kid friendly.}\end{tabular}                   &                                                                                                                                                         \\ 
\hline
\begin{tabular}[c]{@{}c@{}}Low-resource\\fine-tuning\end{tabular}     & \textcolor[rgb]{0.984,0.737,0.016}{The Vaults is a moderately priced} \textcolor[rgb]{0.918,0.263,0.208}{French restaurant} \textcolor[rgb]{0.275,0.741,0.776}{located near Raja Indian Cuisine.}                                                                                                                                                                                                      & 2 IMs                                                                                                                                                   \\ 
\hline
\begin{tabular}[c]{@{}c@{}}Low-resource\\cycle-training\end{tabular}  & \begin{tabular}[c]{@{}l@{}}\textcolor[rgb]{0.259,0.522,0.957}{The Vaults is a French restaurant} \textcolor[rgb]{0.275,0.741,0.776}{near Raja Indian Cuisine} \textcolor[rgb]{0.204,0.659,0.325}{in the riverside area. }\textcolor[rgb]{1,0.427,0.004}{It is kid}\\\textcolor[rgb]{1,0.427,0.004}{friendly} \textcolor[rgb]{0.984,0.737,0.016}{and moderately priced.}\end{tabular}                   &                                                                                                                                                         \\ 
\hline
\multicolumn{3}{|c|}{\textbf{DART - Sample 2}}                                                                                                                                                                                                                                                                                                                                                                                                                                                                                                                                                  \\ 
\hline
\textbf{Input Triple Set}    & \multicolumn{2}{l|}{\begin{tabular}[c]{@{}l@{}}1. \textcolor[rgb]{0.259,0.522,0.957}{[S] K-2 1000 m [P] silver [O] Bulgaria Berenike Faldum Daniela Nedeva}\\2. \textcolor[rgb]{0.918,0.263,0.208}{[S] K-2 1000 m [P] gold [O] Germany Anne Knorr Debora Niche}\\3. \textcolor[rgb]{0.984,0.737,0.016}{[S] K-2 1000 m [P] bronze [O] Hungary Aliz Sarudi Erika Medveczky}\end{tabular}}                                                                                                                                                                          \\ 
\hline
\textbf{Reference Text}      & \multicolumn{2}{l|}{- Anne Knorr won gold, Berenike Faldum won silver, and Aliz Satudi won bronze.}                                                                                                                                                                                                                                                                                                                                                                                                                                                              \\ 
\hline
\textbf{Method}              & \multicolumn{1}{c|}{\textbf{Generation}}                                                                                                                                                                                                                                                                                                                                                               & \textbf{Remark}                                                                                                                                         \\ 
\hline
\begin{tabular}[c]{@{}c@{}}Fully-supervised\\fine-tuning\end{tabular} & \begin{tabular}[c]{@{}l@{}}\textcolor[rgb]{0.259,0.522,0.957}{Bulgaria, Berenike Faldum, Daniela Nedeva won silver,} \textcolor[rgb]{0.918,0.263,0.208}{Germany Anne Knorr, Debora}\\\textcolor[rgb]{0.918,0.263,0.208}{Niche won gold,} \textcolor[rgb]{0.984,0.737,0.016}{Hungary Aliz Sarudi, Erika Medveczky won bronze.}\end{tabular}                                                             &                                                                                                                                                         \\ 
\hline
\begin{tabular}[c]{@{}c@{}}Low-resource\\fine-tuning\end{tabular}     & Bulgaria Berenike Faldum Daniela Nedeva won gold in the K-2 1000 m.                                                                                                                                                                                                                                                                                                                                    & \begin{tabular}[c]{@{}c@{}}2 IMs\\1 FE\end{tabular}                                                                                                     \\ 
\hline
\begin{tabular}[c]{@{}c@{}}Low-resource\\cycle-training\end{tabular}  & \begin{tabular}[c]{@{}l@{}}\textcolor[rgb]{0.259,0.522,0.957}{Bulgaria Berenike Faldum Daniela Nedeva won silver,} \textcolor[rgb]{0.918,0.263,0.208}{Germany Anne Knorr Debora Niche}\\\textcolor[rgb]{0.918,0.263,0.208}{won gold,} \textcolor[rgb]{0.984,0.737,0.016}{Hungary Aliz Sarudi and Erika Medveczky won bronze in K-2 1000 m.}\end{tabular}                                               &                                                                                                                                                         \\ 
\hline
\multicolumn{3}{|c|}{\textbf{DART - Sample 3}}                                                                                                                                                                                                                                                                                                                                                                                                                                                                                                                                                                                                                                                                                                 \\ 
\hline
\textbf{Input Triple Set}                                             & \multicolumn{2}{l|}{\begin{tabular}[c]{@{}l@{}}1. \textcolor[rgb]{0.259,0.522,0.957}{[S] Illinois 2 [P] result [O] Lost re-election Republican gain}\\2. \textcolor[rgb]{0.918,0.263,0.208}{[S] Barratt O'Hara [P] first elected [O] 1948}\\3. \textcolor[rgb]{0.984,0.737,0.016}{[S] Illinois 2 [P] candidates [O] Richard B. Vail R 53.6\% Barratt O'Hara D 46.4\%}\\4. \textcolor[rgb]{0.204,0.659,0.325}{[S] Illinois 2 [P] incumbent [O] Barratt O'Hara}\\5. \textcolor[rgb]{1,0.427,0.004}{[S] Barratt O'Hara [P] party [O] Democratic}\end{tabular}}                                                                                                        \\ 
\hline
\textbf{Reference Text}                                               & \multicolumn{2}{l|}{- Incumbent barratt o'hara was first elected in 1948.0.}                                                                                                                                                                                                                                                                                                                                                                                                                                                                                                                                                                                       \\ 
\hline
\textbf{Method}                                                       & \multicolumn{1}{c|}{\textbf{Generation}}                                                                                                                                                                                                                                                                                                                                                        & \textbf{Remark}                                                                                                                                                                                                                                              \\ 
\hline
\begin{tabular}[c]{@{}c@{}}Fully-supervised\\fine-tuning\end{tabular} & \textcolor[rgb]{0.918,0.263,0.208}{Barratt O'Hara was first elected in 1948.0.}                                                                                                                                                                                                                                                                                                                 & 4 IMs                                                                                                                                                                                                                                                            \\ 
\hline
\begin{tabular}[c]{@{}c@{}}Low-resource\\fine-tuning\end{tabular}     & \textcolor[rgb]{0.918,0.263,0.208}{Barratt O'Hara was first elected in 1948.}                                                                                                                                                                                                                                                                                                                   & 4 IMs                                                                                                                                                                                                                                                            \\ 
\hline
\begin{tabular}[c]{@{}c@{}}Low-resource\\cycle-training\end{tabular}  & \textcolor[rgb]{0.259,0.522,0.957}{In Illinois 2, Barratt O'Hara lost re-election Republican gain }\textcolor[rgb]{0.918,0.263,0.208}{and was first elected in 1948.}                                                                                                                                                                                                                           & 3 IMs                                                                                                                                                                                                                                                            \\ 
\hline
\multicolumn{3}{|c|}{\textbf{DART - Sample 4}}                                                                                                                                                                                                                                                                                                                                                                                                                                                                                                                                                  \\ 
\hline
\textbf{Input Triple Set}    & \multicolumn{2}{l|}{\begin{tabular}[c]{@{}l@{}}1. \textcolor[rgb]{0.259,0.522,0.957}{[S] Clowns [P] eat type [O] pub}\\2. \textcolor[rgb]{0.918,0.263,0.208}{[S] Clowns [P] price range [O] more than £30}\\3. \textcolor[rgb]{0.984,0.737,0.016}{[S] Clowns [P] customer rating [O] 3 out of 5}\\4. \textcolor[rgb]{0.204,0.659,0.325}{[S] Clowns [P] near [O] All Bar One}\end{tabular}}                                                                                                                                                                       \\ 
\hline
\textbf{Reference Text}      & \multicolumn{2}{l|}{- For a 3 out of 5 star customer rated pub near All Bar One, Clowns is relatively expensive.}                                                                                                                                                                                                                                                                                                                                                                                                                                                \\ 
\hline
\textbf{Method}              & \multicolumn{1}{c|}{\textbf{Generation}}                                                                                                                                                                                                                                                                                                                                                               & \textbf{Remark}                                                                                                                                         \\ 
\hline
\begin{tabular}[c]{@{}c@{}}Fully-supervised\\fine-tuning\end{tabular} & \begin{tabular}[c]{@{}l@{}}\textcolor[rgb]{0.259,0.522,0.957}{Clowns is a pub}\textcolor[rgb]{0.204,0.659,0.325}{ near All Bar One.} \textcolor[rgb]{0.984,0.737,0.016}{It has a customer rating of 3 out of 5} \textcolor[rgb]{0.918,0.263,0.208}{and a price range}\\\textcolor[rgb]{0.918,0.263,0.208}{of more than £30.}\end{tabular}                                                              &                                                                                                                                                         \\ 
\hline
\begin{tabular}[c]{@{}c@{}}Low-resource\\fine-tuning\end{tabular}     & \textcolor[rgb]{0.259,0.522,0.957}{Clowns is a pub} \textcolor[rgb]{0.984,0.737,0.016}{with a customer rating of 3 out of 5}                                                                                                                                                                                                                                                                           & 2 IMs                                                                                                                                                   \\ 
\hline
\begin{tabular}[c]{@{}c@{}}Low-resource\\cycle-training\end{tabular}  & \begin{tabular}[c]{@{}l@{}}\textcolor[rgb]{0.259,0.522,0.957}{Clowns is a pub} \textcolor[rgb]{0.204,0.659,0.325}{near All Bar One.} \textcolor[rgb]{0.984,0.737,0.016}{It has a customer rating of 3 out of 5} \textcolor[rgb]{0.918,0.263,0.208}{and a price range}\\\textcolor[rgb]{0.918,0.263,0.208}{of more than £30.}\end{tabular}                                                              &                                                                                                                                                         \\ 
\hline
\end{tabular}
\caption{Error analysis of additional data samples generated by various cycle training settings, for the DART dataset.}
\label{tab:appendix_table2}
\end{table*}

\end{document}